\documentclass[journal]{IEEEtran}
\usepackage{algorithm}
\usepackage{algorithmic}
\usepackage{graphicx}
\usepackage{cite}
\usepackage{makecell}
\usepackage{array}
\usepackage{multirow}
\usepackage{amsmath}
\usepackage{amssymb}

\newcommand\relphantom[1]{\mathrel{\phantom{#1}}}
\begin{document}
%
\title{Active Transfer Learning Network: A Unified Deep Joint Spectral-Spatial Feature Learning Model For Hyperspectral Image Classification}

\author{Cheng~Deng,~\IEEEmembership{Member,~IEEE,}
        Yumeng~Xue,
        Xianglong~Liu,
        Chao~Li,
        Dacheng~Tao,~\IEEEmembership{Fellow,~IEEE}
\thanks{Manuscript received April 19, 2005; revised August 26, 2015. This work was supported in part by the National Natural Science Foundation of China under Grant 61572388 and 61872021, in part by the Key R\&D Program-The Key Industry Innovation Chain of Shaanxi under Grant 2017ZDCXL-GY-05-04-02, and in part by Australian Research Council Projects FL-170100117, DP-180103424, and IH180100002.}
\thanks{Y. Xue, C. Deng, and C. Li are with the School of Electronic Engineering, Xidian University, Xi'an 710071, China (e-mail :chdeng.xd@gmail.com).}
\thanks{X. Liu is with the State Key Lab of Software Development Environment, Beihang University, Beijing 100191, China (email: xlliu@nlsde.buaa.edu.cn).}
\thanks{D. Tao is with the UBTECH Sydney Artificial Intelligence Centre and the School of Information Technologies, the Faculty of Engineering and Information Technologies, the University of Sydney, 6 Cleveland St, Darlington, NSW 2008, Australia (email: dacheng.tao@sydney.edu.au). }
\thanks{\copyright 20XX IEEE. Personal use of this material is permitted. Permission from IEEE must be obtained for all other uses, in any current or future media, including	reprinting/republishing this material for advertising or promotional purposes, creating new collective works, for resale or redistribution to servers or lists, or reuse of any copyrighted component of this work in other works.}}

\markboth{}
{Shell \MakeLowercase{\textit{et al.}}: Bare Demo of IEEEtran.cls for IEEE Journals}

\maketitle

\begin{abstract}
Deep learning has recently attracted significant attention in the field of hyperspectral images (HSIs) classification. However, the construction of an efficient deep neural network (DNN) mostly relies on a large number of labeled samples being available. To address this problem, this paper proposes a unified deep network, combined with active transfer learning that can be well-trained for HSIs classification using only minimally labeled training data. More specifically, deep joint spectral-spatial feature is first extracted through hierarchical stacked sparse autoencoder (SSAE) networks. Active transfer learning is then exploited to transfer the pre-trained SSAE network and the limited training samples from the source domain to the target domain, where the SSAE network is subsequently fine-tuned using the limited labeled samples selected from both source and target domain by corresponding active learning strategies. The advantages of our proposed method are threefold: 1) the network can be effectively trained using only limited labeled samples with the help of novel active learning strategies; 2) the network is flexible and scalable enough to function across various transfer situations, including cross-dataset and intra-image; 3) the learned deep joint spectral-spatial feature representation is more generic and robust than many joint spectral-spatial feature representation. Extensive comparative evaluations demonstrate that our proposed method significantly outperforms many state-of-the-art approaches, including both traditional and deep network-based methods, on three popular datasets.
\end{abstract}

\begin{IEEEkeywords}
Deep learning, hyperspectral image classification, multiple feature representation, active learning, stacked sparse autoencoder (SSAE), transfer learning.
\end{IEEEkeywords}

\IEEEpeerreviewmaketitle

\section{Introduction}
\IEEEPARstart{O} {wing} to the rapid development of remote sensing technology, hundreds of nearly continuous spectral bands and an enormous amount of spatial information can be captured simultaneously via the hyperspectral sensors. Hyperspectral images (HSIs) have been widely utilized in diverse fields, such as precision agriculture \cite{1}, geological exploration \cite{2}, and environmental sciences \cite{3}\cite{4}, where land cover classes usually need to be identified using a small set of training samples through HSIs classification. However, some unfavorable factors also exist that can seriously decrease the classification accuracy when high dimensional spectral/spatial features are involved: 1) the so-called “curse of dimensionality”, in which high-dimensional spectral information hinders the extraction of available
spectral properties; 2) inadequate use of the spectral and spatial information, which significantly restrains the performance of the classifier; 3) the limited availability of labeled samples makes it hard for the classifier to learn the distribution of the HSIs data completely.

Significant efforts have been made to solve these problems using a range of different approaches, which can be divided into three main categories: supervised classification method, unsupervised classification method, and semi-supervised classification method. In recent decades, some typical supervised classification algorithms including SVM \cite{5} \cite{65}, k-nearest-neighbors, and logistic regression \cite{66} have been investigated for HSIs. In \cite{5}, the kernel based SVM was first proposed to learn the feature representation of spectral bands. Besides, instead of using the full spectral bands for data processing, transformation \cite{6} \cite{7}, principal component analysis (PCA) \cite{9,60,61} and other unsupervised dimensionality reduction methods have been exploited to interpret the relevance of spectral bands. Furthermore, the semi-supervised classification methods for HSIs are provided with some available labeled data in addition with unlabeled data. Bruzzone \emph{et al.} \cite{68} proposed a semi-supervised transductive SVM to maximize the hyperplane between the labeled and the unlabeled samples simultaneously. In \cite{69}, a new semi-supervised HSIs classification algorithm is proposed to exploit both hard and soft labels for better modeling the phenomenon of mixed pixels present in HSIs.

Above mentioned methods almost only focus on the spectral bands in HSIs. In order to improving the interpretation of HSIs, many approaches have incorporated spectral feature with abundant spatial contextual information \cite{16,17,57}, e.g., extended morphological profiles (EMPs) \cite{18}, nonnegative matrix factorization \cite{55} \cite{63}, and conditional random field \cite{54}. However, these traditional low-level features are more sensitive to local changes occurred in input data, which greatly reduces the classification accuracy.

Deep neural network (DNN) has been proven to be able to automatically learn a hierarchical feature representation, which is more robust to HSIs classification \cite{13,14,64}. This type of feature is invariant to local changes and thus more suitable for handling the variable spectral/spatial signatures in HSIs \cite{14}. Deep belief network (DBN) \cite{24}, convolutional neural network (CNN) \cite{25,59,67} and stacked autoencoder (SAE) \cite{26,27,28} have been introduced into HSIs classification and have improved classification performance significantly. DBN is usually combined with PCA in order to learn a joint spectral-spatial feature for HSIs classification \cite{31} \cite{32}, while CNN-based HSIs classification models tend to learn spatial information with 2-D patches \cite{33}. In \cite{26}, SAE is first proposed to learn deep feature representation from stacked spectral-spatial feature. Tao \emph{et al.} \cite{28} exploited the stacked sparse autoencoder (SSAE) to extract deep sparse feature representation for HSIs classification, where deep multi-scale spatial features are learned from various patch sizes. In fact, DNN requires a large number of training samples to learn the parameters in different layers. Unfortunately, only a limited number of labeled samples are available for HSIs in practice, and labeling pixels manually is quite time-consuming.

Active learning (AL) and transfer learning (TL) are two popular techniques that have been employed to promote the training process by selecting some unlabeled data for labeling or by using knowledge obtained from related data. AL is an iterative procedure that involves selecting a small set of the most informative unlabeled samples with a query function to train a robust classifier. The training procedure utilizing active sampling data performs more efficiently, because these samples are more suitable for describing the distribution of the unlabeled data. AL for HSI classification was studied using a small set of training samples in \cite{38,39,40,41,42,deng2018active}. Some researchers have adopted AL to perform high-quality sample labeling in order to construct a well-trained DNN. Liu \emph{et al.} \cite{51} constructed a DBN using AL and fewer training samples than are typically used in traditional semi-supervised learning methods. Unlike AL, TL aims to propagate useful knowledge from a source domain to a target domain. In recent years, TL has been successfully applied in the remote sensing field~\cite{43,44,45,46,47} and has coped well with the variability of spectral/spatial information in related HSIs that are acquired by the same sensor at different time or locations . In \cite{45}, TL is combined with CNN to train the target data using auxiliary source data and a limited number of target samples.

Inspired by the ideas of AL and TL, this paper first presents a deep joint spectral-spatial feature representation model that incorporates with active transfer learning for HSI classification. Compared to the shallow methods, the novel deep joint spectral-spatial feature learning framework shows that the learned deep spectral-spatial feature representation is more discriminative for HSI classification and avoids designing the artificial parameters that are sensitive to the local changes of the input data, especially when the training data are limited. More specifically, in contrast to traditional feature extraction methods that stack the original spectral feature with spatial neighborhood information directly, we utilize a hierarchical SSAE network to learn a deep joint spectral-spatial feature that can effectively discover the underlying contextual and structure information in HSIs, which takes full advantage of the variable spectral and spatial features. Such hierarchical SSAE network contains much less parameters than CNN and it is pre-trained with limited labeled samples selected through the AL strategy on source domain. Moreover, active transfer learning is exploited to transfer the pre-trained SSAE network and few training data from the source domain to the target domain, where two different AL strategies are utilized: one selects a limited number of the most informative samples from target domain, while the other removes those samples that are incompatible with the target distribution from the source domain respectively. The pre-trained SSAE network is then fine-tuned with these few updated training samples, which greatly enhances the efficiency of the training process and makes the network more flexible for related HSIs classification tasks. In this way, a generic and robust feature representation is obtained with few high quality samples in our active transfer learning network. Extensive experiments over three widely-used datasets demonstrate that our proposed network, which is based on active transfer learning, has powerful transfer capability under various situations and significantly outperforms several state-of-the-art methods in terms of classification performance.

The rest of this paper is organized as follows. In Section \ref{sec:ssae}, both the sparse AE and stacked sparse AE models are introduced in detail. Section \ref{sec:frame} introduces the framework of our proposed methods. Experiments and analysis are presented in Section \ref{sec:exp}. Section \ref{sec:con} concludes of this paper.

\section{Sparse Autoencoder Model}\label{sec:ssae}
In this section, we briefly introduce a robust model of SSAE, which is adopted to learn a sparse and discriminative feature representation for HSI classification.

\subsection{Sparse Autoencoder}
\begin{figure}[!h]
\centering
\includegraphics[width=2.22in]{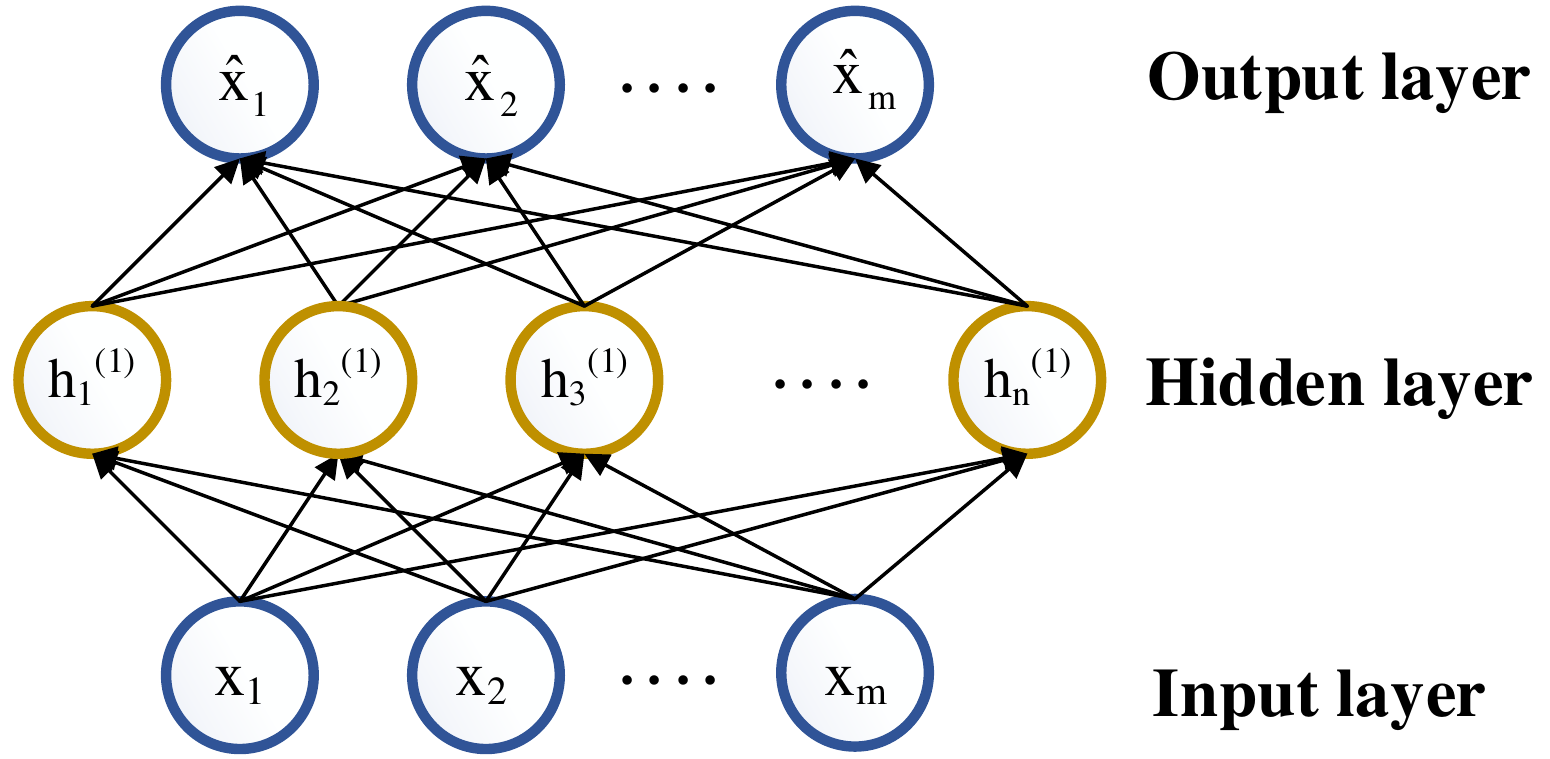}
\caption{The architecture of a shallow sparse autoencoder, which contains an input layer, a hidden layer and an output layer.}
\label{fig:AE}
\end{figure}

 AE is constructed using three layers, \emph{i.e.,} an input layer, a hidden layer, and a reconstruction layer (output layer). From the input to the hidden layer, AE first maps the input $x \in \mathbb{R}^m$ to the hidden layer and generates a latent representation $h \in \mathbb{R}^n$, a step termed the "encoder" step. Feature $ \hat x \in \mathbb{R}^m$ is then decoded from the hidden layer to the reconstruction layer, which is regarded as an abstract representation of the input data. Fig.~\ref{fig:AE} clearly shows the relationship among these three layers.

We formulate the two steps (i.e. ``encoder'' and ``decoder'') as follows:
\begin{equation}
h=S({ W }_{ h }x+{ b }_{ h }),
\end{equation}
\begin{equation}
{\hat x}=S({ W }_{ \hat x }h+{ b }_{ \hat x }),
\end{equation}
where $W_h$ and $W_{\hat x}$ denote the ``encoder'' and the ``decoder'' weights respectively, while $ b_h$ and $b_{\hat x}$ represent the biases of the hidden and reconstruction units.

The activation function $S(\cdot )$, which is used to calculate the value of units in different layers, is generally set to be a sigmoid function. It can be formulated as:
\begin{equation}
S(x)=\frac { 1 }{ 1+{ e }^{ -x } } .
\end{equation}

In fact, the aim of AE is to learn an approximation from the identity function that makes the reconstruction data as similar as possible to the input. Therefore, a loss function is designed to measure the difference between the input data and the reconstruction data, which can be described as:
\begin{equation}
\begin{aligned}
{ J }(W,b)&=\frac { 1 }{ m } \sum _{ i=1 }^{ m }{ \left( \frac { 1 }{ 2 } { \left\| { h_{W,b} }({ x }^{ (i) })-{\hat x}^{ (i) } \right\|  }^{ 2 } \right)  }\\
&\relphantom{=} {}+\frac { \lambda  }{ 2 } \sum _{ l=1 }^{ { n }_{ i }-1 }{ \sum _{ i=1 }^{ { s }_{ i } }{ \sum _{ j=1 }^{ { s }_{ i+1 } }{ { ({ W }_{ ji }^{ (l) }) }^{ 2 } },}  }
\end{aligned}
\end{equation}
where
\begin{equation}
\begin{aligned}
 h_{W,b} ({ x }^{ (i) })=S({ W }{ x ^{(i)}}+b).
\end{aligned}
\end{equation}
The first term of the loss function uses a $\ell_2$ norm to measure the difference between the input data and the reconstruction data, while the second term of the loss function is a regulation term used to prevent over-fitting, and $\lambda$ is a weight decay parameter balancing the effect of these two terms.

Unlike a simple autoencoder model, which learns a low-level compressed representation of the input, SAE \cite{28} can learn an ``overcomplete'' representation by setting the number of hidden units to be larger than the number of input units.

In order to enforce some hidden units inactive for most of the time in SAE, the sparse parameter ${\rho}$ (which is close to zero) \cite{53} is constrained by the average activation ${ \hat { \rho  }  }_{ j }$ of a hidden unit. An extra sparse penalty term is added to the loss function to punish the average activations far away from ${\rho}$. The loss function of the SAE\cite{56} can be rewritten as:
\begin{equation}
{ J }_{ sparse }(W,b)={ J(W,b) }+\beta \sum _{ j=1 }^{ { s }_{ 2 } }{ KL(\rho \left\| { \hat { \rho  }  }_{ j } \right)  },
\end{equation}
where
\begin{equation}
\rm KL(\rho \left\| { \hat { \rho  }  }_{ j } \right) =\rho \log\frac{\rho}{{ \hat { \rho  }  }_{ j }}+(1-\rho )\log\frac{1-\rho}{1-{ \hat { \rho  }  }_{ j }}.
\end{equation}
Here, parameter $\beta $ is the weight of the sparse penalty, while ${ \rm KL(\rho \left\| { \hat { \rho  }  }_{ j } \right)  } $ is the Kullback-Leibler (KL) divergence used to measure the difference between mean ${ \rho  }$ and mean ${ \hat { \rho  }  }_{ j }$. The KL value increases rapidly when the difference between ${ \hat { \rho  }  }_{ j }$ and ${ \rho  }$ grows. Thus, the sparse penalty term enforces ${ \hat { \rho  }  }_{ j }$ close to ${ \rho  }$ when this value reaches the minimum.

\begin{figure}[!t]
\centering
\includegraphics[width=3.55in]{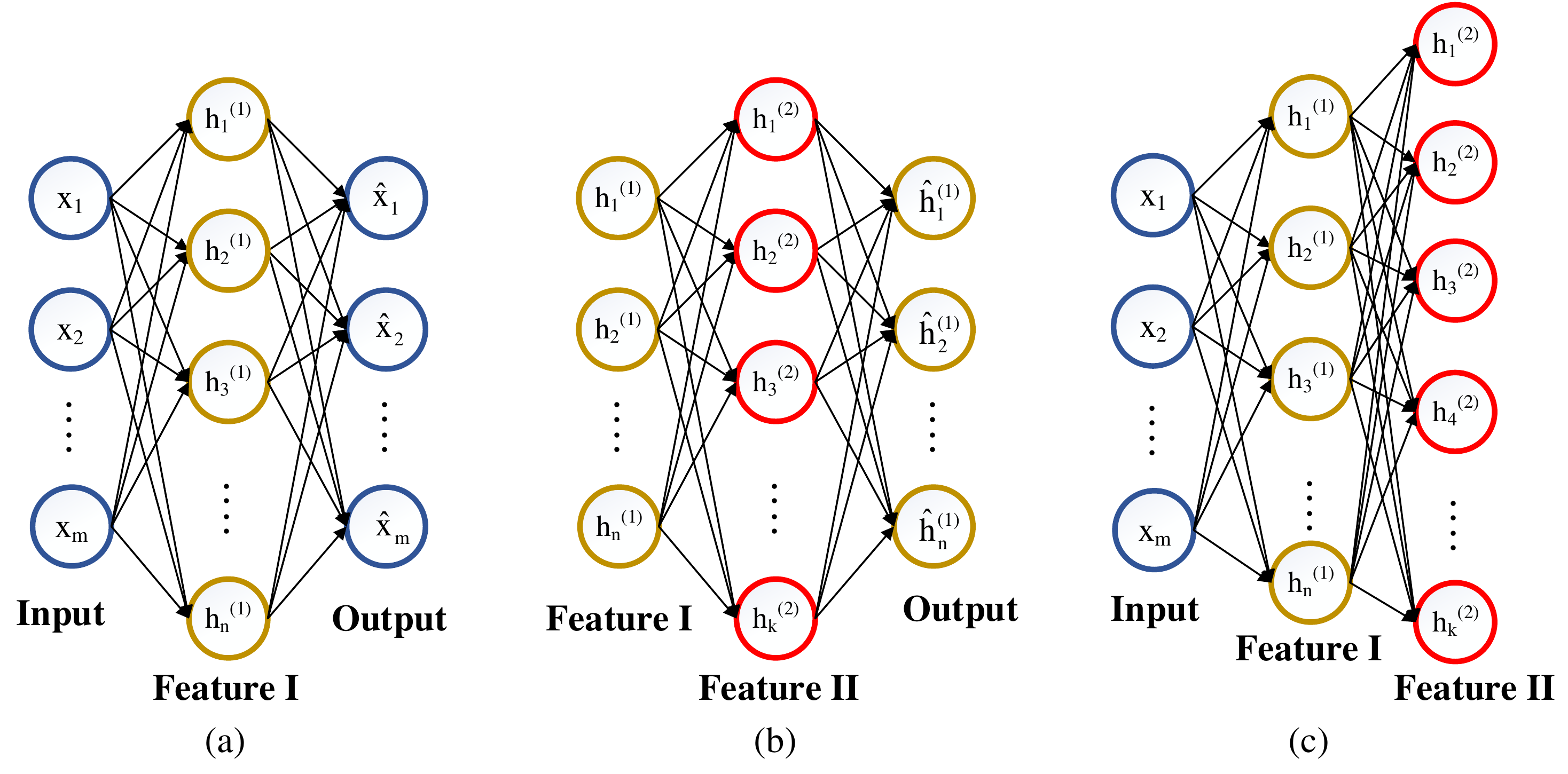}
\caption{The construction of a stacked sparse autoencoder with two hidden layers. (a) sparse autoencoder1. (b) sparse autoencoder2. (c) SSAE which
consists of two previous basic shallow sparse autoencoder.}
\label{fig:SSAE}
\end{figure}

\subsection{Stacked Sparse Autoencoder (SSAE)}
SSAE \cite{28} is a deep architecture of SAE that stacks several hidden layers of basic SAE together, meaning that the output of each layer is regarded as the input of the subsequent layer in SSAE.

Fig.~\ref{fig:SSAE} illustrates the construction process of SSAE specifically. First, the original input data \textbf x is trained by a SAE for learning a nonlinear feature ${\mathbf h }^{ (1) }={ S }_{ 1 }(\mathbf x)$. This feature is regarded as the input of the secondary SAE and used to extract the more abstract feature ${\mathbf h }^{ (2) }={ S }_{ 2 }(\mathbf{ h }^{ (1) })$. Finally, a SSAE with two hidden layers is formed by stacking these two SAEs together, allowing the learning of a deep feature from the input with a transform function $\mathbf{ h }^{ (2) }={ S }_{ 2 }({ S }_{ 1 }(\mathbf{ x }^{ (1) }))$. More technical details of SSAE can be found in \cite{53} and \cite{56}.

\section{Proposed Method}
\label{sec:frame}
In previous work, traditional deep learning models for HSI classification have tended to learn the high-level feature representation of the stacked spectral-spatial features \cite{28}\cite{34}. These models are thus unable to take full advantage of the information contained in HSIs. Moreover, it is very difficult to obtain a large amount of labeled data to construct a well-trained DNN, and the learned feature representation is only suitable for a specific data set. These problems have motivated us to develop a novel hierarchical SSAE network that incorporates with active transfer learning, enabling the learning of a generic and robust spectral-spatial feature representation.

\begin{figure*}[!t]
\centering
\includegraphics[width=18cm]{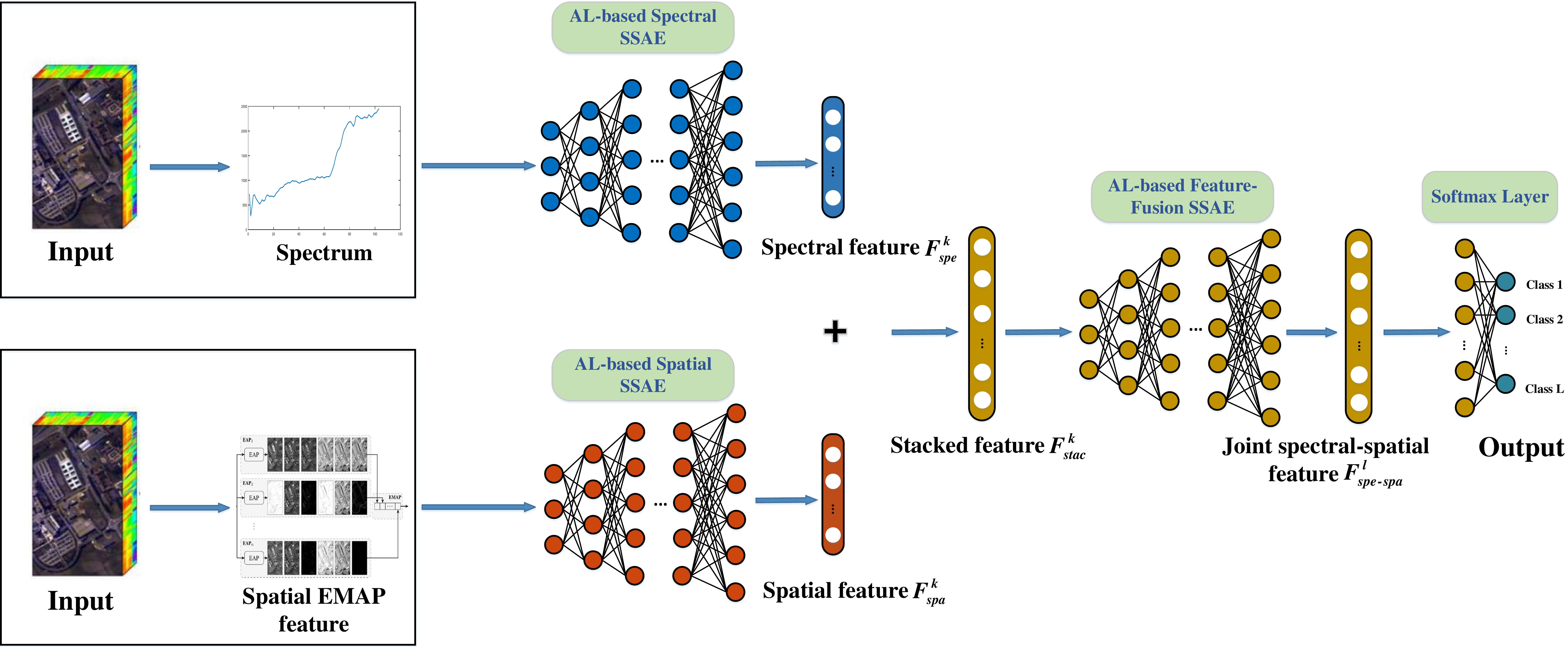}
\caption{The complete architecture of proposed deep multiple feature fusion model, combined with three AL based SSAEs for extracting and fusing the deep spectral and spatial features.}
\label{fig:framework}
\end{figure*}

\subsection{Architecture of the Proposed Method }
Fig.~\ref{fig:framework} shows the architecture of our proposed method. Spectral and spatial features are first extracted by two different, well-designed sub-networks. The two obtained features are then stacked together and input into the sub-network that follows. In this way, a joint deep spectral-spatial feature that is more suitable for our classification task can be achieved.

From spectral perspective, an SSAE with \emph{k} hidden layers is designed to extract a more abstract spectral feature ${ F }_{ spe }^{ k } $ for a given HSI. Unlike the spectral characteristic, spatial information is represented as a 2-D structure, which contains a lot of useful contextual information for our classification task. If we directly exploit an end-to-end deep learning model to extract spatial feature, the spatial neighborhood information should be extracted within small patches, and then vectorized to 1-D structure, which may lead to the loss of the contextual information \cite{33}. Moreover, such end-to-end learning model generally depends on a great quantity of training samples with strong supervision. Inspired by \cite{16}, an extended morphological attribute profile (EMAP) is utilized in our preprocessing stage to maintain the spatial structure information in 1-D structure as the input of SSAE without any supervision, which superiority can be shown in Section IV. The EMAP can learn the raw spatial structure information from the first principal components of HSI by using multiple attribute profiles (APs). The shallow model proposed in \cite{20} chose the output of EMAP as a spatial feature and combined it with the original spectral feature. In our framework, spatial information is preprocessed by EMAP with standard deviation and area attributes, then sent into a SSAE branch as a 1-D signal to learn its corresponding deep spatial feature ${ F }_{ spa }^{ k } $.

We have now obtained two deep descriptors, ${ F }_{ spe }^{ k } $  and ${ F }_{ spa }^{ k } $, in our feature extraction sub-networks. Next, we design a task-driven fusion method, which stacks the learned deep spectral and spatial feature (i.e., ${ F }_{ spe }^{ k } $  and ${F}_{spa}^{k}$) together as a new feature ${F}_{stac}^{k}$ before feeding it into another SSAE. This SSAE uses $l$ hidden layers to fuse the stacked spectral-spatial feature, after which the last hidden layer outputs a feature that can be regarded as the deep joint spectral-spatial feature ${F}_{spe-spa}^{k}$. This feature contains most of spectral and spatial information in HSI. Finally, the deep joint spectral-spatial feature is classified by the softmax regression layer, which is used to predict the conditional probability distribution of each class as follows:
\begin{equation}
P({ y=i }|\mathbf{x},W,b)=\frac { { e }^{ { W }_{ i }\mathbf{x}+{ b }_{ i } } }{ \sum _{ i=1 }^{ k }{ { e }^{ { W }_{ j }\mathbf{x}+{ b }_{ j } } }  },
\end{equation}
where $W$ and $b$ are the weight and bias of the softmax regression layer respectively.

\subsection{Active Sampling Strategy for Pre-trained Network}
In order to pre-train hierarchical SSAE networks on the source domain and achieve an impressive performance, a large amount of labeled training samples are required for supervised learning. However, only a very limited amount of labeled samples are available in practice for training SSAE in the task of HSI classification, which is prone to over-fitting. To address this problem, the batch-mode AL method is considered here for use in selecting a small set of high-quality samples so that SSAE can be trained in a more effective way.

In the AL strategy, there are various query functions (or criteria) can be chosen for sample selection, including margin sampling (MS) \cite{38} and multiclass-level uncertainty (MCLU) \cite{39}. In most previous work, AL has generally been applied to the shallow models like multi-class SVM \cite{58}. An SAE with AL procedure was proposed for HSI classification in \cite{41}, where the uncertainty criterion of the query function depends on the output of the softmax layer and the most uncertain samples are added iteratively into the training set in order to retrain the classifier.

\begin{figure}[!t]
\centering
\includegraphics[width=0.45\textwidth]{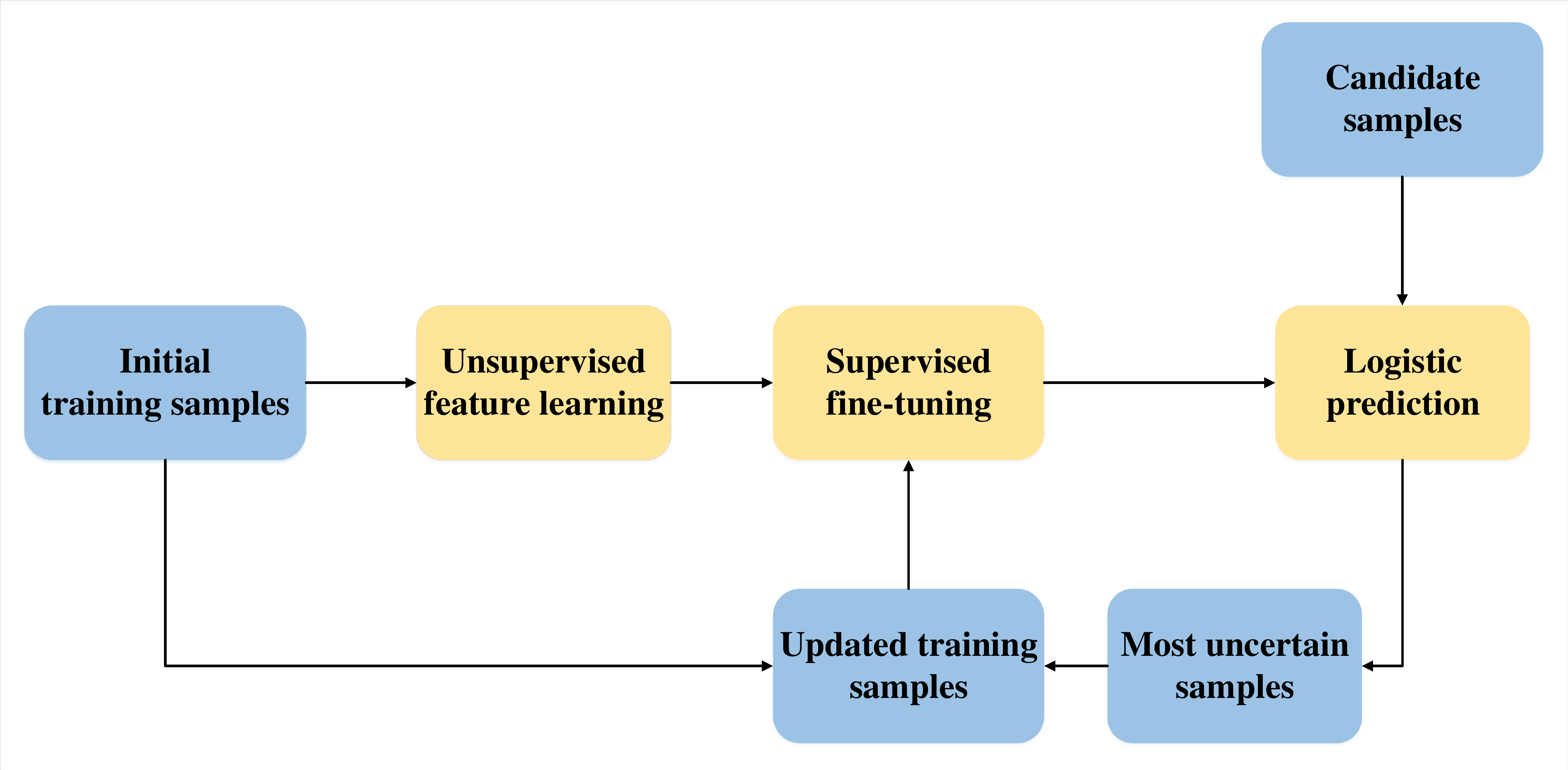}
\caption{The proposed batch-mode AL sampling strategy for SSAE.}
\label{fig:AL}
\end{figure}

The flowchart of our proposed batch-mode AL sampling method is shown in Fig.~\ref{fig:AL}. As illustrated, we first exploit the greedy layer-wise training strategy to train SSAE layer by layer with a few training samples, then the learned features of these samples and their corresponding labels are used to train a softmax classifier and fine-tune SSAE with the softmax layer by using backward propagation. These training samples are sufficient to train SSAE to learn a favorable feature representation, which can be verified in Section IV. Next, a subset of unlabeled data regarded as the candidate set is classified with softmax regression. Finally, AL iteratively selects the most uncertain unlabeled samples, adds them into the training set with true labels, and simultaneously removes them from the candidate set. Unlike the AL sampling procedure reported in \cite{41}, our proposed method first exploits a few training samples and their corresponding labels to train and fine-tune SSAE with a softmax layer, and then in order to boost the performance of SSAE, AL selects some most informative samples to refine-tune SSAE with the softmax layer at each iteration, instead of only retraining the parameters the softmax layer. By this way, a well-trained SSAE can be obtained more effectively with limited training samples.

Here, the MCLU technique is chosen as the query criterion, which applies a difference function ${ c }_{ diff }(\mathbf {x})$ to record the uncertainty of unlabeled samples. ${ c }_{ diff }(\mathbf {x})$ on logistic regression \cite{42} considers the difference between largest and second largest class-conditional-probability density as the following object function:
\begin{equation}\label{eqn:8}
{ c }_{ diff }(\mathbf {x})= { p }^{ (i) }(\mathbf {x}\left| {\omega  }_{ max1 } \right) -{ p }^{ (i) }(\mathbf {x}\left| { \omega  }_{ max2 } \right),
\end{equation}
where
\begin{equation}
{\omega  }_{ max1 }=\mathop{\arg\max} \limits_{{ \omega  }_{ n }\in \Omega }\{ { p }^{ (i) }(\mathbf {x}|{ \omega  }_{ n })\},
\end{equation}
\begin{equation}
\hspace{3mm}{ \omega  }_{ max2 }=\mathop{\arg\max} \limits_{ { \omega  }_{m }\in \Omega /\left\{ { \omega  }_{ max1 } \right\}  }\{ { p }^{ (i) }(\mathbf {x}|{ \omega  }_{ m })\}.
\end{equation}

If the value of ${ c }_{ diff }(\mathbf {x})$ is large, the possibility that $x$ belongs to $\omega _{max1}$ is high. On the contrary, a small ${ c }_{ diff }(\mathbf {x})$ indicates that $x$ will be assigned to the predicted class $\omega _{max1}$ with a low confidence. Under these circumstances, $x$ is treated as an uncertain sample and should be manually labeled in order to better describe the distribution of the feature space. Namely, MCLU technique selects a subset of unlabeled samples with the minimum value of ${ c }_{ diff }(\mathbf {x})$; this subset contains more information about the unlabeled data to be labeled by the supervisor.

\subsection{Active Knowledge and Samples Transfer Learning}
To facilitate the learning of a generic and flexible feature representation for multiple related HSIs and improve the performance of target task with limited training samples, we here propose an active transfer learning method which tranfers the knowledge and training samples learned from the source domain to the target domain.

Traditional TL methods have been successful in the high-resolution remote sensing images (HRRS) classification task. They usually prefer to transfer the bottom layers of a network that has been pre-trained on ImageNet \cite{43}\cite{45} or on the related images \cite{46} \cite{47} to the target network, and then employ target data to train the top layers of target network for HRRS classification. In our proposed active transfer learning framework, we not only adopt the network pre-trained on source HSIs, but also pay more attention to the relationship between the distribution of source and target HSI data. The technical details are introduced in Algorithm \ref{alg:A}. We firstly initialize hierarchical SSAE networks using the training samples on the source domain by an AL sampling strategy. The pre-trained network and training set on the source domain are then transferred to the target domain simultaneously. Subsequently, we iteratively update the training set with two different criteria on the target and source domain (\emph{i.e.}, the sample selecting criterion in Eq.~\eqref{eqn:11} and the sample removing criterion in Eq.~\eqref{eqn:13}), which are used to fine-tune the pre-trained SSAE network.
\begin{equation}
\mathbf {h^+}=argmin\{ { p }^{ (i) }(\mathbf {x}\left| {\omega  }_{ max1 } \right) -{ p }^{ (i) }(\mathbf {x}\left| { \omega  }_{ max2 } \right) \}.
\label{eqn:11}
\end{equation}

\begin{algorithm}[!t]
\caption{The proposed active transfer learning method}
\label{alg:A}
\begin{algorithmic}
\REQUIRE
\STATE{$s^-$: the number of samples removed from the training set }
\STATE\hspace*{0.25in}{on the source domian at each iteration; }
\STATE{$t^+$: the number of unlabeled samples needing to be queried } \STATE\hspace*{0.25in}{from the target data at each iteration; }
\STATE{$T_s$: the training set on source data for initializing SSAE; }
\STATE{$U_t$: pool of target unlabeled samples needing to be}
\STATE\hspace*{0.25in}{queried and labeled in the softmax layer. }

\STATE\rule[0.05em]{8.5cm}{0.05em}
\end{algorithmic}

\begin{algorithmic}[1]
\STATE {Initialize the AL based SSAE with source training samples $T ^{(0)}=T_s$.}
\STATE {Transfer the pre-trained network and training set $T ^{(0)}$ to $U_t$.}
\STATE {Learn the deep feature representation of unlabeled data in $U_t$ and classify these data in the softmax layer.}
\REPEAT
\STATE Select the set $S^+$ of $t^+$ samples from $U_t$ using sample selecting criterion in Eq.~\eqref{eqn:11}.
\STATE Remove the set $S^-$ of $s^-$ samples from $T_s$ based on sample removing criterion in Eq.~\eqref{eqn:13}.
\STATE The selected samples $S^+$ are assigned a label by the supervisor.
\STATE Update $T^{(i+1)}=\{{T^{(i)}/S^-}\}\bigcup { S^+ } $.
\STATE Fine tune all the SSAE parameters with the updated training set $T^{(i+1)}$.

\UNTIL{a stopping criterion is satisfied.}
\end{algorithmic}
\end{algorithm}

The AL with MCLU sampling strategy shown in Eq.~\eqref{eqn:8} iteratively selects the minimum number of the most informative samples from the target domain and add them to the source training set $T_s$. Meanwhile, the training samples in $T_s$ that can not match the distribution of the updated training set in ${i}$-th iteration will be discarded. The criterion for removing source domain data is as follows:
\begin{equation}
{ c }_{ rem }(\mathbf {x})= { p }^{ (0) }(\mathbf {x}|{ \omega  }_{ t })-{ p }^{ (i) }(\mathbf {x}|{ \omega  }_{ t }),
\end{equation}
where ${ c }_{ rem }(\mathbf {x})$ measures the difference between class-conditional-probability densities of source training samples and the updated training data obtained at the $i$-th iteration for each sample $x\in T_s$. If this value is small, the distribution of the class $\omega _t$ changes only a little for source data $x$. On the other hand, a large ${ c }_{ rem }(\mathbf {x})$ indicates that the distribution of class $\omega _t$ on the source domain has shifted gradually towards target data after ${i}$ iterations and $x$ can no longer describe the distribution of the target domain. Therefore, $x$ should be removed from the training set. Sample removing criterion selects the sample $\mathbf {h^-}$ that has a maximum value of ${ c }_{ rem }(\mathbf {x})$, as follows:
\begin{equation}\label{eqn:13}
\mathbf {h^-}= \mathop{\arg\max}\limits_{ x\in { T }^{ (0) } }\{ { p }^{ (0) }(\mathbf {x}|{ \omega  }_{ t })-{ p }^{ (i) }(\mathbf {x}|{ \omega  }_{ t })\}.
\end{equation}

Finally, the training set updated by the two criteria iteratively fine-tunes the source pre-trained network, gradually tailoring it to the distribution of target domain until the stopping criterion is satisfied. Meanwhile, it learns a generic and robust feature representation for the target data.

The stopping criterion is determined by the value of the loss function of SSAE, which guarantees the convergence of the algorithm. As we know, during the training process of SSAE, the value of the loss function first decreases fast, and then oscillates near the minimum and the value of classification accuracy reaches its best at the meantime. Thus, the iteration of active transfer learning will stop if the value of the loss function is less than $\varepsilon $, where $\varepsilon $ is defined according to the experiments.

\section{Experiments and Discussions}\label{sec:exp}
In this section, in order to comprehensively evaluate the performance of our proposed method, we carry out different experiments on three popular hyperspectral data sets and compare our method with several state-of-the-art alternatives. Moreover, all the following experiment results are generated on a Windows 7 personal computer equipped with a 64-bit Intel Core i5-3470 CPU running at 3.2GHz and an 8GB RAM. All the proposed methods are implemented with MATLAB R2015b.
\begin{figure*}
\centering
\includegraphics[width=18cm]{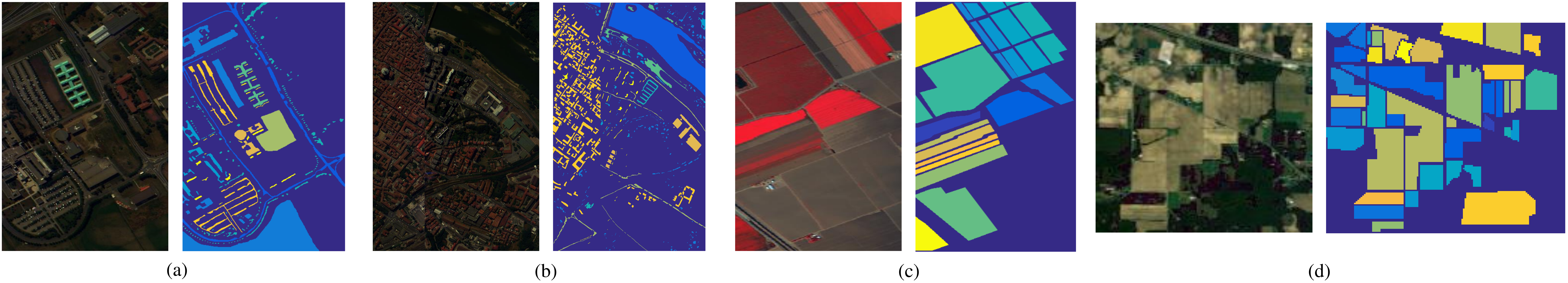}
\caption{The false color maps (Left) and groundtruth maps (Right) of four datasets. (a) Pavia University data (bands 58, 29, and 3). (b) Pavia Center data (bands 60, 30, and 2). (c) Salinas Valley data (bands 52, 25, and 10). (d) Indian Pines data (Bands 25, 19, and 8).}
\label{fig:graph5}
\end{figure*}

\subsection{Datasets and Settings}

We employ four widely-used hyperspectral datasets in the experiments: \emph{Pavia University}, \emph{Pavia Center}, \emph{Salinas Valley} and \emph{Indian Pines}. Fig. \ref{fig:graph5} shows their false color map and groundtruth map respectively. Besides, Tables \ref{tab:tab1} and \ref{tab:tab2} also show some details about the three data sets.

\subsubsection{Pavia University}
This dataset was collected by the Reflective Optics System Imaging Spectrometer (ROSIS) sensor over Pavia, Italy. 103 useful spectral bands remain after removing noise-affected bands from the 115 total bands. The size of the image on a band is $610\times340$ pixels. There are 9 classes of land cover used in the experiment. Fig. \ref{fig:graph5} (a) shows the false
color map and groundtruth map of Pavia University.

\subsubsection{Pavia Center}
 This dataset was obtained in the same way as \emph{Pavia University}. There are 9 different types of land objects in total. The image size of \emph{Pavia Center} is $1096\times715\times102$ pixels after deleting a 381-pixel-wide black band. The false color map and the groundtruth map are shown in Fig. \ref{fig:graph5} (b).

\subsubsection{Salinas Valley}
This dataset was acquired by the Airborne Visible/Infrared Imaging Spectrometer (AVIRIS) sensor over Salinas, California, and contains 204 bands after removing water absorption bands and noisy bands. The image size is 512 x 217 pixels, and there are 16 classes of land covers on the scene. The false color image and groundtruth map are shown in Fig. \ref{fig:graph5} (c).

\begin{table}[!t]
\centering  
\caption{Number of Reference Data in Each Class\protect\\for Pavia University and Pavia Center}
\renewcommand\arraystretch{1}
\begin{tabular}{p{54pt}|c|p{54pt}|cp{54pt}cp{54pt}c|} 
\hline
\hline
\multicolumn{2}{c|}{\centering Pavia University} &\multicolumn{2}{c}{\centering Pavia Center}\\
\hline
\centering Class &Reference data &\centering Class &Reference data\\ \hline  
\centering Asphalt  &6631 &\centering Water  &65971\\         
\centering Meadows  &18649  &\centering Trees &7598\\        
\centering Gravel &2099 &\centering Asphalt &3090\\
\centering Trees &3064 &\centering Bricks &2685 \\
\centering Metal Sheets &1345 &\centering Bitumen &6584 \\
\centering Bare Soil &5029 &\centering Tiles &9248 \\
\centering Bitumen &1330 &\centering Shadows &7287\\
\centering Bricks &3682 &\centering Meadows &42826 \\
\centering Shadows  &947 &\centering Bare Soil &2863 \\\hline
\centering Total  &42776 &\centering Total &148152 \\ \hline
\hline
\end{tabular}
\label{tab:tab1}
\end{table}

\begin{table}[!t]
\centering  
\caption{Number of Reference Data in Each Class\protect\\for Salinas Valley}
\renewcommand\arraystretch{1.0}
\begin{tabular}{p{54pt}|cp{54pt}c} 
\hline
\hline
\multicolumn{2}{c}{\centering Salinas Valley} \\
\hline
\centering Class &Reference data\\ \hline  
\centering Greenweeds1  &2009 \\         
\centering Greenweeds2  &3726  \\        
\centering Fallow &1976\\
\centering Rough Fallow &1394  \\
\centering Smooth Fallow &2678  \\
\centering Stubble &3959 \\
\centering Celery &3579  \\
\centering Grapes &11271  \\
\centering Vinyard Soil &6203  \\
\centering Corn &3278  \\
\centering Lettuce 4wk &1068  \\
\centering Lettuce 5wk &1927  \\
\centering Lettuce 6wk &916  \\
\centering Lettuce 7wk &1070  \\
\centering Untrain vinyard &7268  \\
\centering Vertical vinyard  &1807  \\\hline
\centering Total  &54129  \\ \hline
\hline
\end{tabular}
\label{tab:tab2}
\end{table}

All of these four datasets are divided into three parts: the training, candidate and test sets. We first disorder the data in each class randomly, and then select $r$ ($r$=25, 50, 75, 100) labeled samples from each class for training, 20\% of the unlabeled data per class as the candidate data and the rest data of the image data for testing. Moreover, in order to avoid the randomness in sampling and the marginal performance, we repeat the experiments 10 times and use the mean value of the classification results to evaluate the performance of the proposed methods.

In active transfer learning experiment, we consider various transfer situations, including cross-dataset and intra-image transfer. In cross-dataset case, \emph{Pavia University} and \emph{Pavia Center} can be used as source datasets for each other because they have some classes of land covers in common. As for \emph{Salinas Valley}, we apply \emph{Indian Pines} (collected by the same sensor with \emph{Salinas Valley}) to pre-train hierarchical SSAE networks. \emph{Indian Pines} contains 16 classes of land covers, but their types differ from those in \emph{Salinas Valley}. The details can be found in Fig. \ref{fig:graph5} (d). We assume that the four datasets all have sufficient samples for training. For the intra-image case, we select two different regions as source and target domains respectively among three datasets respectively; these regions have the same classes of land covers.

As for the evaluation metric, commonly used statistics such as overall accuracy (OA), average accuracy (AA), and Kappa coefficient (Kappa) \cite{48} are applied to record and assess the performance of different classification methods.

\subsection{SSAE Structure Analysis}
Unlike the shallow feature extraction models, SSAE can learn the feature distribution automatically. Thus, the setting of the structure parameters for SSAE will play an important role in the quality of the extracted features. Here, we investigate how the number of hidden layers in SSAE and different query functions of AL influence the classification performance.

\begin{table}[t]
\centering  
\caption{The Sensitivity of Feature-Extraction SSAE \protect\\over The Number of Hidden Layers}
\renewcommand\arraystretch{1.25}
\begin{tabular}{m{45pt}|c|m{50pt}|cm{40pt}|c|m{50pt}c}  
\hline
\hline
\centering Number of hidden layers & Pavia University &\centering Pavia Center &Salinas Valley\\
\hline
\centering 1  &98.59\% &\centering 99.34\%  &92.52\%\\         
\centering 2  &\centering \textbf{98.78}\%  &\centering 99.46\% &\textbf{94.60}\% \\  
\centering 3 &98.75\% &\centering \textbf{99.49}\% &94.24\%\\ \hline
\hline
\end{tabular}
\label{tab:tab3}
\end{table}

\begin{table}[!t]
\centering  
\caption{The Sensitivity of Feature-Fusion SSAE \protect\\over The Number of Hidden Layers}
\renewcommand\arraystretch{1.25}
\begin{tabular}{m{45pt}|c|m{50pt}|cm{40pt}|c|m{50pt}c}  
\hline
\hline
\centering Number of hidden layers & Pavia University &\centering Pavia Center &Salinas Valley\\
\hline
\centering 1  &99.33\% &\centering 99.63\%  &92.69\%\\         
\centering 2  &\centering \textbf{99.35}\%  &\centering 99.65\% &\textbf{95.13}\% \\  
\centering 3 &99.13\% &\centering \textbf{99.69}\% &94.45\%\\  \hline
\hline
\end{tabular}
\label{tab:tab4}
\end{table}

\subsubsection{Depth Effect}

The number of hidden layers in SSAE is the key factor affecting the classification performance, which determines the abstraction level of the input features. Here, we change only the number of hidden layers and fix other parameters to check the corresponding performance. For each dataset, we use 3\% labeled samples per class to train the network. The weights and biases in each layer are initialized randomly and optimized by minimizing the loss function in a greedy layer-wise manner. We try several SSAEs with different depths vary from 1 to 3 layers. In this experiment, it is suggested that the units in the first hidden layer are set to learn an ``overcomplete'' feature representation of the input data during the unsupervised feature learning stage. For \emph{Pavia University} and \emph{Pavia Center}, the classification performance reaches its best value when the number of units for each hidden layer in feature-extraction SSAE is set to 200, 150, and 100, and the number of units for each hidden layer in feature-fusion SSAE is set to 400, 200, and 150. For \emph{Salinas Valley}, each hidden layer in feature-extraction SSAE is composed of 400, 300, 200 units and each hidden layer in feature-fusion SSAE is composed of 800, 400, 300 units. The number of output units in the softmax layer is equal to the number of classes of land covers in different datasets. The training samples are used for pre-training and fine-tuning the whole SSAE with the softmax layer. The number of iterations for both the unsupervised training and supervised fine-tuning stages is set to 500. maximal permitted number of iterations of softmax classifier is 200. For the standard stochastic gradient descent method, the sparsity parameter is set to 0.1, while the weight decay parameter \cite{49} is fixed as 7$\times { 10 }^{ -7 }$ and the learning rate is 0.05. The sparsity penalty weight is set to 0.05.

The depth effect on the classification results for different datasets is shown in Tables \ref{tab:tab3} and \ref{tab:tab4}. Table \ref{tab:tab3} illustrates the sensitivity of feature-extraction SSAE over the number of hidden layers, while Table \ref{tab:tab4} evaluates the sensitivity of feature-fusion SSAE. From the classification results shown in these tables, we can see that when the number of hidden layers changes from 1 to 2, OA increases significantly. However, these values flatten out when the depth changes from 2 to 3. For \emph{Salinas Valley}, the classification performance can be seen to decline by 0.68\% in Table \ref{tab:tab4} when the hidden layer number increases. Therefore, in our proposed network, we construct SSAE with two hidden layers for both feature extraction and feature fusion.

\subsubsection{Query Function Effect}
\begin{figure*}
\centering
\includegraphics[width=18cm]{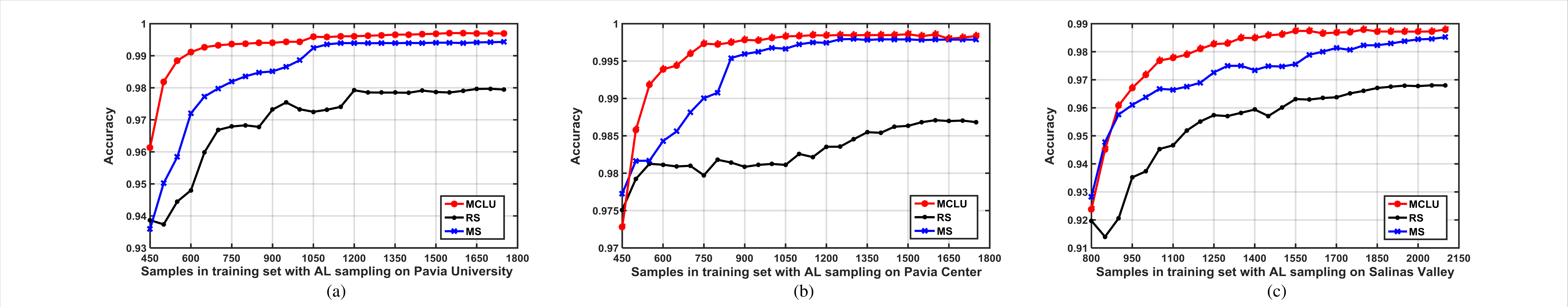}
\caption{The AL curves with different query functions of 26 iterations on three datasets. The number of query samples is 50. (a) Pavia University. (b) Pavia Center. (c) Salinas Valley. }
\label{fig:graph6}
\end{figure*}

In order to verify the effectiveness of AL sampling with MCLU method, we compare MCLU technique with other two typical query functions: namely random sampling method and MS method. In experiments, we first divide the dataset into 3 parts: 50 labeled samples of each class are randomly selected for training SSAE, and 20\% of the unlabeled data per class is regarded as the candidate set. The rest of the reference data are exploited for use in testing the classification performance. The training samples are used to pre-train the parameters of SSAE with two hidden layers. Three different query functions are then combined with softmax layer respectively to select 50 most uncertain samples (query step) from the candidate set at each iteration. These samples are added into the training set with true labels to fine-tune the network and simultaneously removed from the candidate set. For the setting of superior limit of label queries (the budget \emph{N}), we use 26 active learning iterations with 1,300 samples to check the changing trend of the classification results. We then record the classification results of AL based SSAE with three different query functions, as shown in Fig. \ref{fig:graph6}. Comparing the classification results among three datasets, we find that the MCLU technique always outperforms the other two techniques during the AL procedure on all datasets, especially on \emph{Salinas Valley}. Therefore, the MCLU technique is applied as the query function in our proposed framework.

\begin{table*}[t]
\centering  
\caption{Overall Accuracy of Different Methods on Pavia University Data.}
\renewcommand\arraystretch{1.25}
\begin{tabular}{c|c|c|c|c|c|c}  
\hline
\hline
\centering Ratio of training samples & Spe-EMAP SVM\cite{21} &\centering JSSAE\cite{26} & Spe-SSAE & \centering EMAP-SSAE & Spe-EMAP SSAE & {Proposed}
\\
\hline
\centering 5\%  &93.66\% &\centering 91.93\% &90.85\% &98.05\% &98.76\% &\textbf{99.29}\%\\         
\centering 10\%  &\centering 94.58\%  &\centering 92.64\% &91.43\% &98.30\% &98.85\% &\textbf{99.52}\%\\  
\centering 15\% &\centering 95.01\%  &\centering 92.83\% &91.58\% &98.43\% &98.94\% &\textbf{99.47}\% \\
\centering 20\% &\centering 95.36\%  &\centering 93.64\% &92.28\% &98.62\% &99.06\% &\textbf{99.54}\%\\  \hline
\hline
\end{tabular}
\label{tab:tab5}
\end{table*}

\begin{figure*}
\centering
\includegraphics[width=18cm]{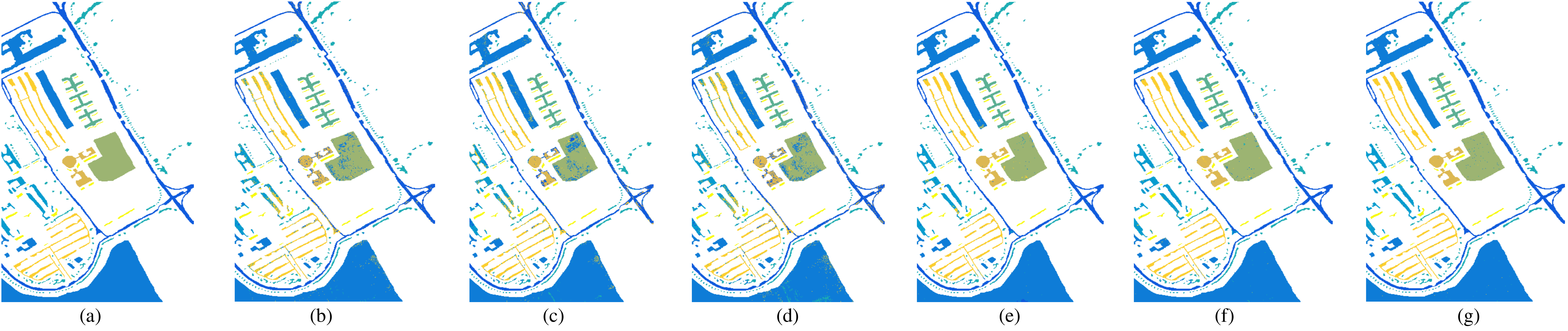}
\caption{Classification maps of different methods on Pavia University data. 5\% of samples per class are chosen for training. (a) Groundtruth. (b) Result
of Spe-EMAP SVM method. (c) Result of JSSAE method. (d) Result of Spe-SSAE method. (e) Result of EMAP-SSAE method. (f) Result of Spe-EMAP SSAE method. (g) Result of
our proposed method.}
\label{fig:graph7}
\end{figure*}

\begin{table*}
\centering  
\caption{Overall Accuracy of Different Methods on Pavia Center Data.}
\renewcommand\arraystretch{1.25}
\begin{tabular}{c|c|c|c|c|c|c}  
\hline
\hline
\centering Ratio of training samples & Spe-EMAP SVM\cite{21} &\centering JSSAE\cite{26} & Spe-SSAE & \centering EMAP-SSAE & Spe-EMAP SSAE & {Proposed}
\\
\hline
\centering 5\%  &98.87\% &\centering 98.77\% &98.38\% &99.27\% &99.42\% &\textbf{99.69}\%\\         
\centering 10\%  &\centering 99.10\%  &\centering 99.04\% &98.55\% &99.34\% &99.69\% &\textbf{99.79}\%\\  
\centering 15\% &\centering 99.23\%  &\centering 99.10\% &98.64\% &99.46\% &99.74\% &\textbf{99.85}\% \\
\centering 20\% &\centering 99.30\%  &\centering 99.16\% &98.73\% &99.62\% &99.76\% &\textbf{99.87}\%\\  \hline
\hline
\end{tabular}
\label{tab:tab6}
\end{table*}

\begin{figure*}
\centering
\includegraphics[width=18cm]{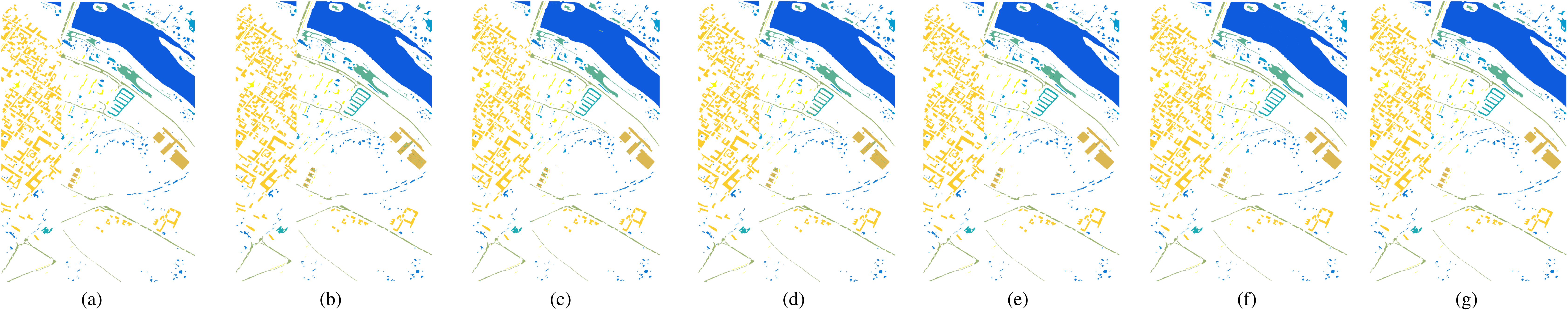}
\caption{Classification maps of different methods on Pavia Center data. 5\% of samples per class are chosen for training. (a) Groundtruth. (b) Result
of Spe-EMAP SVM method. (c) Result of JSSAE method. (d) Result of Spe-SSAE method. (e) Result of EMAP-SSAE method. (f) Result of Spe-EMAP SSAE method. (g) Result of
our proposed method.}
\label{fig:graph8}
\end{figure*}

\begin{table*}
\centering  
\caption{Overall Accuracy of Different Methods on Salinas Valley Data.}
\renewcommand\arraystretch{1.25}
\begin{tabular}{c|c|c|c|c|c|c}  
\hline
\hline
\centering Ratio of training samples & Spe-EMAP SVM\cite{21} &\centering JSSAE\cite{26} & Spe-SSAE & \centering EMAP-SSAE & Spe-EMAP SSAE & {Proposed}
\\
\hline
\centering 5\%  &93.11\% &\centering 90.89\% &89.23\% &96.12\% &96.82\% &\textbf{97.46}\%\\         
\centering 10\%  &\centering 94.07\%  &\centering 91.72\% &89.43\% &96.23\% &96.93\% &\textbf{97.68}\%\\  
\centering 15\% &\centering 94.72\%  &\centering 91.81\% &89.48\% &96.59\% &97.00\% &\textbf{97.90}\% \\
\centering 20\% &\centering 95.11\%  &\centering 92.36\% &89.71\% &96.76\% &97.11\% &\textbf{98.06}\%\\  \hline
\hline
\end{tabular}
\label{tab:tab7}
\end{table*}

\begin{figure*}
\centering
\includegraphics[width=18cm]{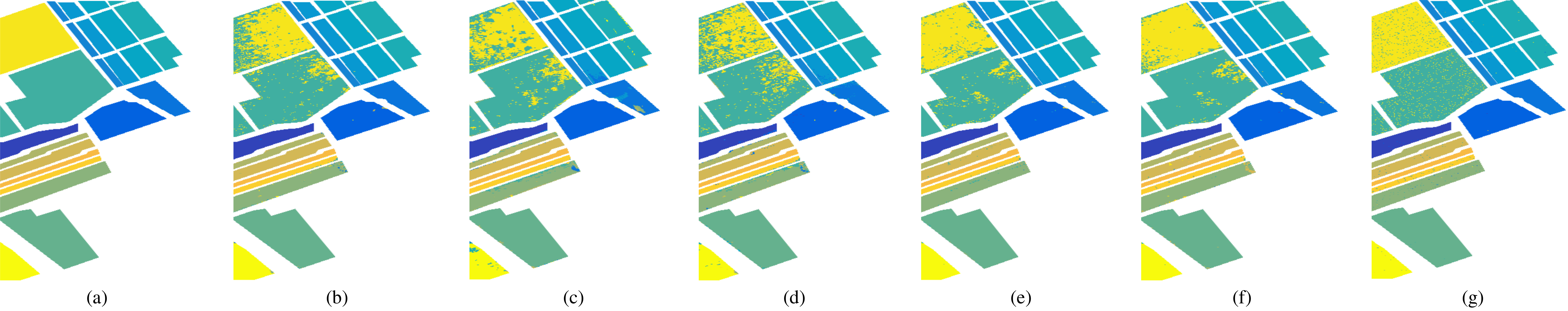}
\caption{Classification maps of different methods on Salinas Valley data. 5\% of samples per class are chosen for training. (a) Groundtruth. (b) Result
of Spe-EMAP SVM method. (c) Result of JSSAE method. (d) Result of Spe-SSAE method. (e) Result of EMAP-SSAE method. (f) Result of Spe-EMAP SSAE method. (g) Result of
our proposed method.}
\label{fig:graph9}
\end{figure*}

\subsection{Comparison with State-of-the-Art Methods}
In this section, we aim to compare our proposed method with five other state-of-the-art classification methods. These methods are: joint spectral-EMAP feature representation with SVM (Spe-EMAP SVM) \cite{21}, joint spectral-spatial feature representation with SAE (JSSAE) \cite{26}, deep spectral feature representation with SSAE (Spe-SSAE), deep EMAP feature representation with SSAE (EMAP-SSAE), and joint spectral-EMAP feature representation with SSAE (Spe-EMAP SSAE). SVM is a typical shallow classification method regarded as a benchmark in the field of HSI classification. Spe-SSAE and EMAP-SSAE are two branches of our proposed method and are used to determine whether or not our deep joint spectral-spatial feature representation model takes effect. The JSSAE method is used to compare the influence of different spatial features on the classification performance.

To set the parameters of these baseline methods, we first tune the parameters for their best performance. The Lib-SVM toolbox \cite{50} is considered to finish Spe-EMAP SVM method. We use linear SVM as the classifier and find the optimal parameters of SVM by means of five-fold cross validation, with SVM parameter $c$ in the range of [2$^{5}$, 2$^{6}$, 2$^{7}$, 2$^{8}$, 2$^{9}$, 2$^{10}$] and $\gamma$ in the range of [2$^{-5}$, 2$^{-4}$, ... , 2$^{4}$, 2$^{5}$]. In order to design a fair comparison, the structure of JSSAE is kept the same as ours, including the number of hidden layers, training iterations and learning rate. The spatial information is extracted from the spatial neighborhood in JSSAE and the size of spatial neighborhood is set to 5$\times$5 pixels. Spe-EMAP SSAE stacks the original spectral feature with the EMAP feature directly to learn the joint spectral-spatial features. The parameters of the rest of the baseline methods are set as illustrated in Section IV-B.

In experiments, we reproduce all the above methods on four different training datasets: 5\%, 10\%, 15\%, and 20\% of the original data. We repeat all of the above HSI classification methods 10 times. The mean of overall accuracy for the three datasets is recorded in Tables \ref{tab:tab5}, \ref{tab:tab6}, and \ref{tab:tab7}. It can be seen that while SVM takes the same features as the SSAE method, the overall accuracy of SSAE outperforms the SVM method on different training sets. This demonstrates that the deep feature is more stable and robust that can raise the effectiveness of HSI classification. The overall accuracy of SSAE that takes EMAP features as input outperforms the JSSAE methods, indicating that spatial structure information extracted by EMAP is more appropriate for SSAE to learn a spatial feature representation than the spatial neighborhood information. Comparing the Spe-EMAP SSAE with the Spe-SSAE and EMAP-SSAE methods, we find that multiple features learning with a deep model does indeed improve the performance of HSI classification. Particularly, our proposed method outperforms other classification methods with different training sets among three datasets, thus demonstrating that the deep joint spectral-spatial feature makes a genuine difference and takes good advantage of the spectral/spatial signatures. Parts of the different classification maps are shown in Figs. \ref{fig:graph7}, \ref{fig:graph8} and \ref{fig:graph9}. Here, we can see that there are far fewer wrongly-labeled pixels in the map of our method than other methods, especially in \emph{Salinas Valley}.

\subsection{Transferability of Active Transfer Learning Network}
\begin{figure*}[!t]
\centering
\includegraphics[width=18cm]{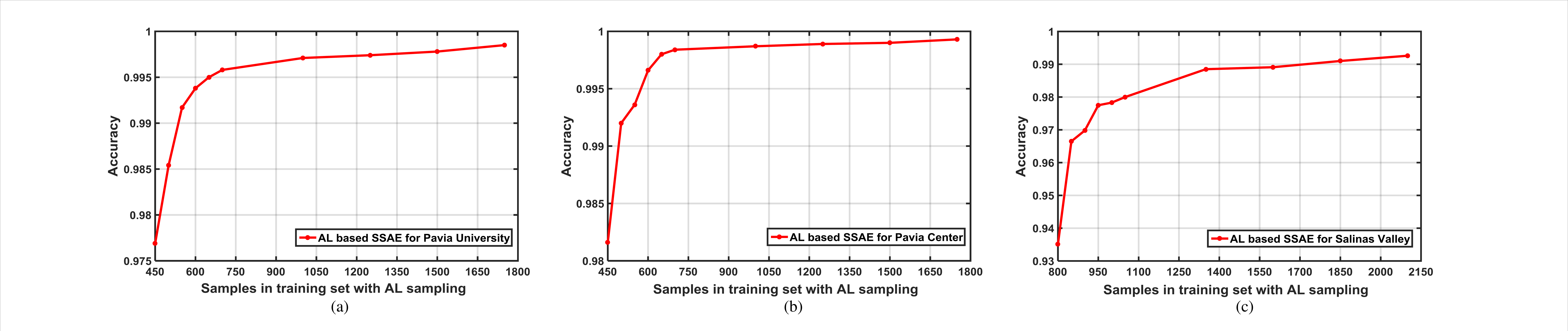}
\caption{Evolution of overall accuracy during 26 AL iterations on three datasets. The number of query samples is 50. (a) Pavia University. (b) Pavia Center. (c) Salinas Valley. }
\label{fig:graph10}
\end{figure*}
\subsubsection{Analysis of AL Procedure of the Pre-trained Network}
In this section, we train our model with the AL sampling strategy over all three datasets. We first set four training sets by randomly selecting 25, 50, 75, and 100 labeled samples in each class. 20\% of unlabeled data per class is regarded as the candidate set. The MCLU technique is exploited as the query function to choose 50 most informative samples from the candidate set, according to the prediction of softmax layer for different classes at each iteration. Finally, the whole network is fine-tuned iteratively by the updated training data. In the experiments, we set the upper limit of AL iterations at 26. The classification results of the AL procedure for the pre-trained network are shown in Fig. \ref{fig:graph10}, and the values of OA, AA, and Kappa in different training sets for three datasets are recorded in Table \ref{tab:tab8}.

As shown in Fig. \ref{fig:graph10}, the curves of AL sampling with 26 iterations demonstrate that the overall accuracy increases rapidlly during the first 4 or 5 iterations, after which the curve becomes flattened. Our method only uses less than half of the labeled samples used in the non-AL method to train SSAE and obtain a promising classification result. The values of OA, AA, and Kappa are all higher than those for the non-AL sampling method shown in Tables \ref{tab:tab5}, \ref{tab:tab6}, and \ref{tab:tab7}, which indicates that the selected most uncertain samples can better describe the distribution of the unlabeled data and effectively avoid labeling the redundant samples.
\begin{table}[!t]
\centering  
\caption{Classification Results of the Proposed AL Procedure for Pre-trained network on Three Data Sets }
\renewcommand\arraystretch{1.25}
\begin{tabular}{c|m{80pt}|c|c|cc}  
\hline
\hline
  &\centering Number of training samples per class  & OA & AA & Kappa \\
\hline
 &\centering 25  &99.78\% &\centering 99.78\%  &0.9970\\         
Pavia &\centering 50  &\centering \textbf{99.85}\%  &\centering \textbf{99.81}\% &\textbf{0.9979} \\  
University &\centering 75 &99.82\% &\centering 99.80\% &0.9975\\
 &\centering 100 &99.80\% &\centering 99.77\% &0.9974\\\hline
\hline
&\centering 25  &99.84\% &\centering 99.49\%  &0.9977\\         
Pavia &\centering 50  &\centering \textbf{99.93}\%  &\centering \textbf{99.81}\% &\textbf{0.9990}\\  
Center &\centering 75 &99.89\% &\centering 99.64\% &0.9984\\
&\centering 100 &99.87\% &\centering 99.60\% &0.9983\\\hline
\hline
&\centering 25  &99.02\% &\centering 99.45\%  &0.9891\\         
Salinas&\centering 50  &\centering \textbf{99.26}\%  &\centering \textbf{99.54}\% &\textbf{0.9918} \\  
Valley&\centering 75 &99.19\% &\centering 99.55\% &0.9912\\
&\centering 100 &99.17\% &\centering 99.48\% &0.9908\\\hline
\hline
\end{tabular}
\label{tab:tab8}
\end{table}


\subsubsection{Transferability of the Pre-trained Network}
In this paper, we adopt the active transfer learning method to learn the various spectral and spatial signatures of different land covers between the source domain and target domain. Here, we transfer four different source training sets and the pre-trained feature representation networks on these source training sets to the target domain. As noted above, \emph{Pavia University} and \emph{Pavia Center} can be used as source data for each other, and the initial training sets on these datasets for pre-training are set at 25, 50, 75, and 100 samples per class. Because there are not enough samples on \emph{Indian Pines}, we use 5\%, 10\%, 15\%, and 20\% labeled data per class to initialize the pre-trained network and then transfer it to \emph{Salinas Valley}.

In order to learn the deep feature representation of spectral and spatial information on the target domain, we first randomly select 20\% of the unlabelled target data as the candidate set; the remaining 80\% of samples are set as the test data. The original source training data are regarded as the initial training set. All the candidate data are then trained by the pre-trained network. AL queries 80 most informative samples in the candidate set and adds them into the training set with the true labels, while 50 samples in the original source training set are iteratively removed from the training set. We extract the deep spectral and spatial feature from the source domain and target domain by the transferred network, after which the deep feature fusion network is initialized by the stacked deep spectral-spatial feature of the original source training data. Finally, the network is transferred by the target deep spectral-spatial feature.

\begin{figure}[!t]
\centering
\includegraphics[width=2.35in]{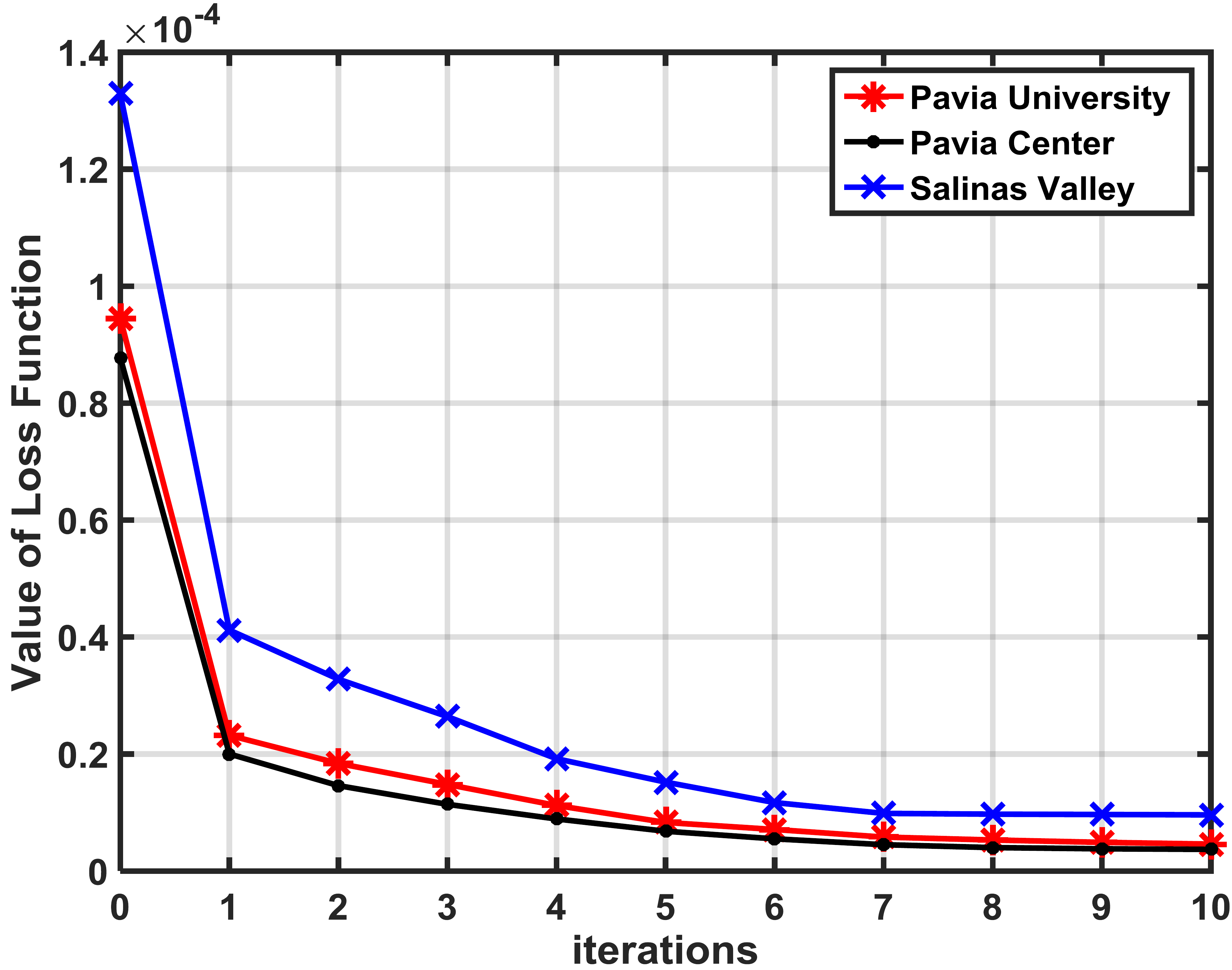}
\caption{Evolution of the loss function over the iterations of active transfer learning.}
\label{fig:graph11}
\end{figure}

During the experiments, we record the value of SSAE loss function after every iteration of active transfer learning in Fig. \ref{fig:graph11} and find that for \emph{Pavia University} and \emph{Pavia Center}, the value of the loss function does not decrease and the value of the classification accuracy reaches its best when the value of the loss function is less than 5$\times$10$^{-6}$, so we set $\varepsilon=5\times10^{-6}$ for these two datasets. As for \emph{Salinas Valley}, we set $\varepsilon=1\times10^{-5}$. Finally, the number of active transfer learning iterations is set to 10 according to the evolution of the loss function on three datasets.

\begin{table}[t]
\centering  
\caption{Classification Results on Pavia University After Transferring \protect\\ the Pre-trained Network and Training samples \protect\\on Pavia Center to Pavia University}
\renewcommand\arraystretch{1.25}
\begin{tabular}{m{80pt}|c|c|cc}  
\hline
\hline
\centering Number of training samples per class  & OA &\centering AA & Kappa \\
\hline
\centering 25  &99.57\% &\centering 99.56\%  &0.9946\\         
\centering 50  &\centering \textbf{99.61}\%  &\centering \textbf{99.61}\% &\textbf{0.9948} \\  
\centering 75 &99.58\% &\centering 99.57\% &0.9947\\
\centering 100 &99.60\% &\centering 99.58\% &0.9948\\\hline
\hline
\end{tabular}
\label{tab:tab9}
\end{table}

\begin{table}[t]
\centering  
\caption{Classification Results on Pavia Center After Transferring \protect\\ the Pre-trained Network and Training samples \protect\\on Pavia University to Pavia Center}
\renewcommand\arraystretch{1.25}
\begin{tabular}{m{80pt}|c|c|cc}  
\hline
\hline
\centering Number of training samples per class  & OA &\centering AA & Kappa \\
\hline
\centering 25  &99.80\% &\centering 99.35\%  &0.9971\\         
\centering 50  &\centering \textbf{99.86}\%  &\centering \textbf{99.62}\% &\textbf{0.9980} \\  
\centering 75 &99.83\% &\centering 99.44\% &0.9976\\
\centering 100 &99.83\% &\centering 99.43\% &0.9975\\\hline
\hline
\end{tabular}
\label{tab:tab10}
\end{table}

\begin{table}[!h]
\centering  
\caption{Classification Results on Salinas Valley After Transferring \protect\\ the Pre-trained Network and Training samples \protect\\on Indian Pines to Salinas Valley}
\renewcommand\arraystretch{1.25}
\begin{tabular}{m{80pt}|c|c|cc}  
\hline
\hline
\centering Ratio of training samples  & OA &\centering AA & Kappa \\
\hline
\centering 5\%  &\centering \textbf{98.61}\%  &\centering \textbf{99.27}\% &\textbf{0.9945} \\         
\centering 10\%  &\centering 98.53\%  &\centering 99.26\% &0.9936\\  
\centering 15\% &98.51\% &\centering 99.09\% &0.9834\\
\centering 20\% &98.51\% &\centering 99.11\% &0.9833\\\hline
\hline
\end{tabular}
\label{tab:tab11}
\end{table}

\begin{figure*}[!t]
\centering
\includegraphics[width=18cm]{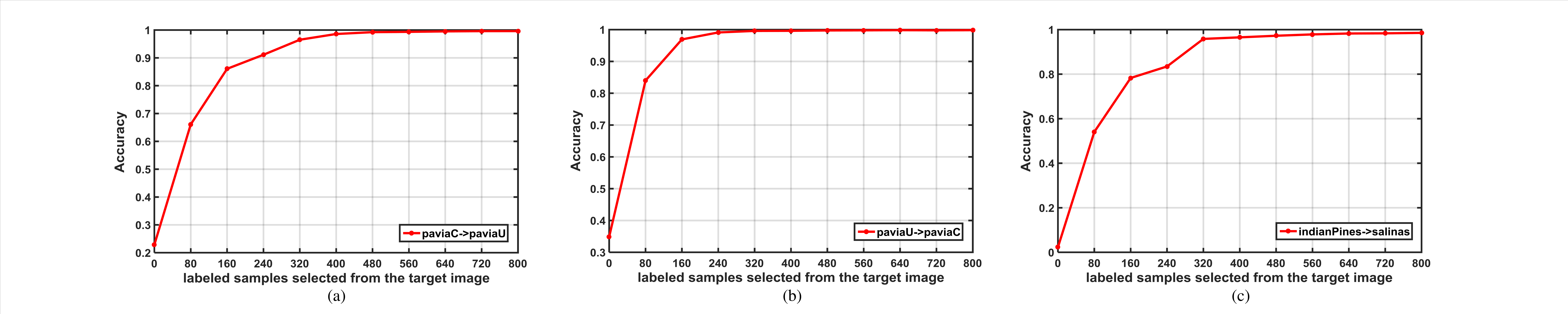}
\caption{Evolution of overall accuracy during 10 transfer learning iterations on three datasets. The number of query samples is 80. (a) Pavia University. (b) Pavia Center. (c) Salinas Valley. }
\label{fig:graph12}
\end{figure*}

The performance results of the active transfer learning methods are presented in Fig. \ref{fig:graph12} and Tables \ref{tab:tab9}, \ref{tab:tab10}, and \ref{tab:tab11}. It is obvious that the active transfer learning method successfully learns the various spectral/spatial signatures in related HSIs and achieves a promising classification result with few target samples. Moreover, it can also be seen that the improvement of the classification performance is influenced by the size of the source training data. When the number of initial training samples per class is more than 50 (on \emph{Pavia University} and \emph{Pavia Center}) or the ratio reaches 10\% (on \emph{Salinas Valley}), the value of overall accuracy slightly declines. This is because more source domain information prevents the network from learning the distribution of the target domain. The curves shown in Fig. \ref{fig:graph12} indicate that while the classification performance for the target domain is very poor when target samples are not used, the overall accuracy increases significantly when only 80 samples are added into the training set. This is due to the effects of the phenomenon of ``domain shift''. Particularly, the disparity between the source domain and the target domain makes a substantial difference to the performance of the active transfer learning method. Fig. \ref{fig:graph12} indicates that our proposed method outperforms on \emph{Pavia University} than on \emph{Salinas Valley}. This is because there are common land cover classes in \emph{Pavia University} and \emph{Pavia Center}, while the classes of land covers between \emph{Indian Pines} and \emph{Salinas Valley} are different.
\begin{table}[!b]
\centering  
\caption{The computational time of the proposed approach}
\renewcommand\arraystretch{1.25}
\begin{tabular}{m{32pt}|m{50pt}|m{35pt}|m{30pt}|c}  
\hline
\hline
\centering &\centering Datasets & \centering Pavia University &\centering Pavia Center &Salinas Valley\\
\hline
\centering Pretrained &\centering training time (min)  &\centering28.62 &\centering 41.20  &62.67\\  
\centering Network&\centering test time (min)  &\centering 0.013  &\centering 0.039 &0.026\\  
\hline
\centering Transferred &\centering training time (min)  &\centering21.68 &\centering 26.51  &54.07\\  
\centering Network&\centering test time (min)  &\centering 0.012  &\centering 0.039 &0.023\\  
\hline
\hline
\end{tabular}
\label{tab:tab12}
\end{table}
\subsubsection{Computational Costs of Active Transfer Learning Network}
We report the computational time of the proposed method in Table \ref{tab:tab12}. In general, when compared with the shallow classification methods, the deep neural network, as the proposed method, takes more time to train the network because it needs iterative calculation. On the other hand, the test time of the deep neural network can be much shorter, which is more important in real classification tasks.
As illustrated in Table \ref{tab:tab12}, we can find that for the pre-trained network on source domain, the training time is respectively 28.62, 41.20, and 62.67 minus for \emph{Pavia University}, \emph{Pavia Center} and \emph{Salinas Valley}, and it only spends 0.013, 0.039 and 0.026 minus testing samples on these datasets, which is much shorter than several deep learning based methods because our proposed method is robust and it can extract the deep joint spectral-spatial feature of test samples quickly. As for the transferred network, the training time is 21.68, 26.51 and 54.07 minus for these three datasets, which is shorter than the training time of the pre-trained network. It is because few target training samples are used to transfer the pre-trained model and the number of active transfer learning iterations is much fewer than the number of AL iterations of the pre-trained model.
\begin{figure*}[!t]
\centering
\includegraphics[width=18cm]{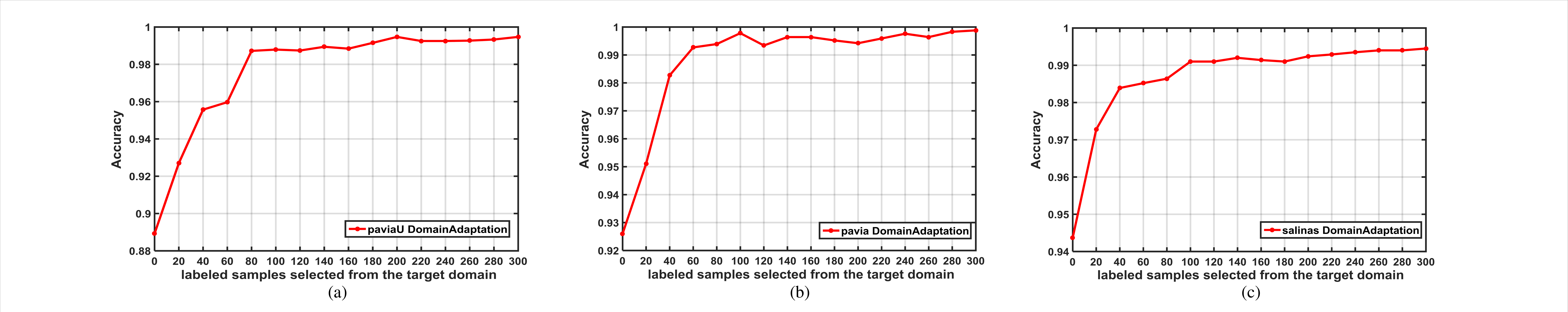}
\caption{Evolution of overall accuracy during 15 transfer learning iterations for domain adaptation on three datasets. The number of query samples is 20. (a) Pavia University. (b) Pavia Center. (c) Salinas Valley. }
\label{fig:graph14}
\end{figure*}

\begin{figure}[!h]
\centering
\includegraphics[width=3.55in]{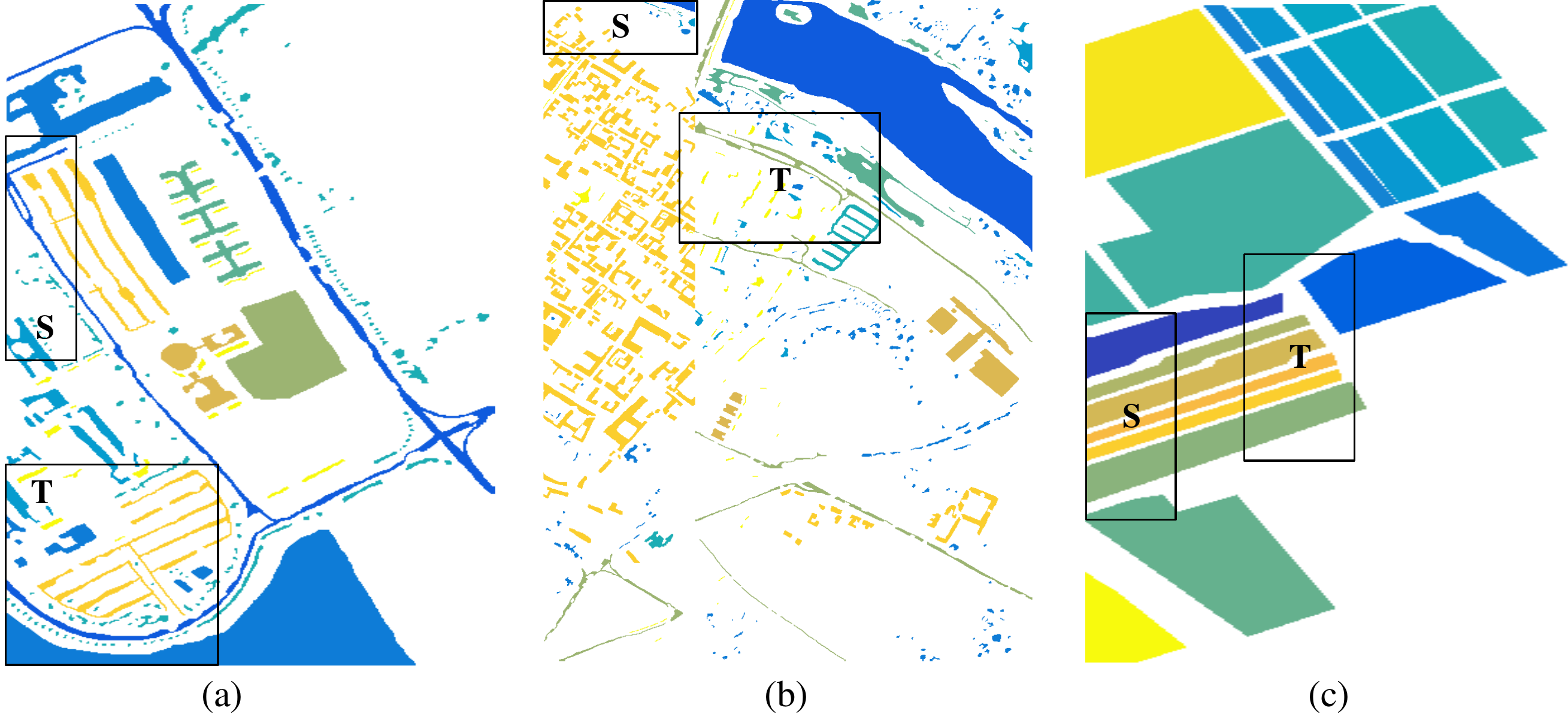}
\caption{Data in source domain and target domain in (a) Pavia University. (b) Pavia Center. (c) Salinas Valley.}
\label{fig:graph13}
\end{figure}

\subsubsection{Domain Adaptation of the Pre-trained Network}

Here, we verify the domain adaptation of the pre-trained network for three datasets. As shown in Fig. \ref{fig:graph13} (a) and (b), the ground-truth maps of the source and target domains contain five classes (class 1, 2, 3, 4, and 8) on \emph{Pavia University} and five classes (class 1, 2, 3, 5, and 8) on \emph{Pavia Center}. On \emph{Salinas Valley}, there are 6 classes (class 1, 10, 11, 12, 13, and 14) shown in Fig. \ref{fig:graph13} (c). For \emph{Pavia University} and \emph{Salinas Valley}, we randomly select 40 samples per class in the source domain to pre-train the network. We use 30 source samples per class in \emph{Pavia Center}. We query 20 samples in the target domain over 15 iterations for all three datasets. The performance of the domain adaptation method is presented in Fig. \ref{fig:graph14} and Table \ref{tab:tab13}. This method relies on very limited training data and still obtains an effective result. We can conclude that active transfer learning does take effect on the domain adaptation, especially when there is a large distribution gap between the source and target domains.
\begin{table}[!h]
\centering  
\caption{Classification Results of the Proposed AL based Domain Adaptation Method on Three Datasets}
\renewcommand\arraystretch{1.25}
\begin{tabular}{c|m{60pt}|c|c|ccc}  
\hline
\hline
\centering Data Set  & \centering training samples in source/target domain &\centering OA &\centering AA & Kappa \\
\hline
\centering Pavia University  &\centering 200/300 &\centering 99.47\%  &99.21\% & 0.9918\\         
\centering Pavia Center  &\centering 150/300  &\centering 99.88\% & 99.68\% &0.9976 \\  
\centering Salinas Valley &\centering 240/300  &\centering 99.45\% & 99.57\% &0.9912\\
\hline
\hline
\end{tabular}
\label{tab:tab13}
\end{table}

\section{Conclusion}\label{sec:con}
In this paper, we propose a novel active transfer learning network for HSI classification, where two SSAE sub-networks are applied to extract deep spectral and spatial features and one sequential SSAE sub-network is used to seamlessly fuse these deep features. The AL sampling method is exploited to select a subset of the most informative unlabeled samples for labeling and add them to the training set at each iteration, which can boost the performance of a pre-trained network with limited labeled samples. Meanwhile, considering the variable spectral/spatial signatures of different land covers in related HSIs, the pre-trained network and the training data from the source domain are transferred to the target domain. Subsequently, the pre-trained network is fine-tuned with the updated training set, which comes from two sources, i.e., the most informative samples in the target domain and the source samples remaining after removing those discrepant with the distribution of the target domain. Experimental results demonstrate that the proposed method exhibits promising performance compared with many state-of-the-art approaches. In the future, the optimal architecture parameters of hierarchical SSAE and the AL sampling criterion used in our method still need further research. Moreover, it is also worth investigating the possibility of exploiting useful transfer knowledge among the data from different sensors.


%

%
%
%
%
%
%
%

\bibliographystyle{ieeetr}
\bibliography{references}

\begin{IEEEbiography}[{\includegraphics[width=1in,height=1.25in,clip,keepaspectratio]{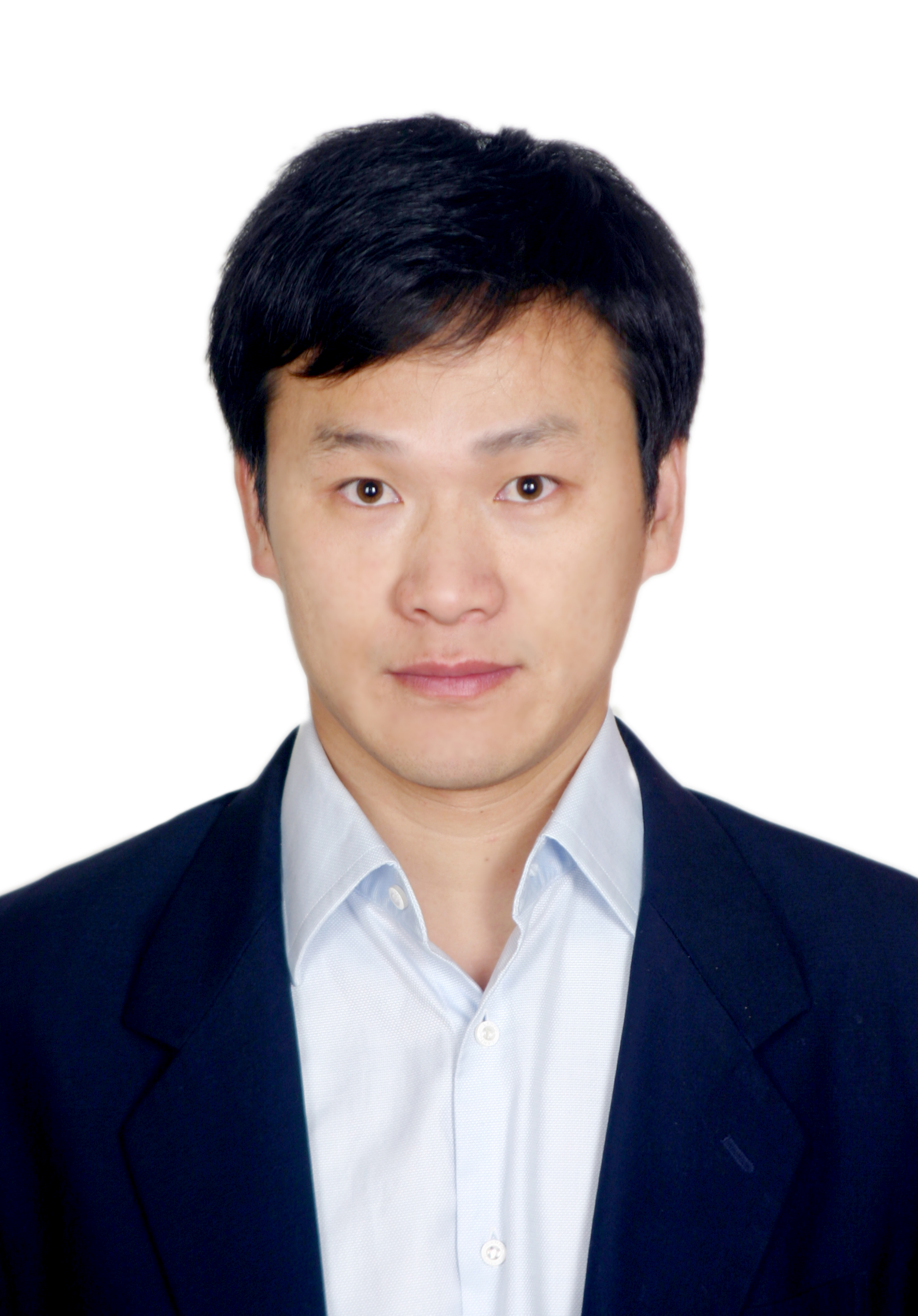}}]{Cheng Deng} (S'09) received the B.E., M.S., and Ph.D. degrees in signal and information processing from Xidian University, Xi'an, China. He is currently a Full Professor with the School of Electronic Engineering at Xidian University. His research interests include computer vision, pattern recognition, and information hiding. He is the author and coauthor of more than 70 scientific articles at top venues, including IEEE T-NNLS, T-IP, T-CYB, T-MM, T-SMC, ICCV, CVPR, ICML, NIPS, IJCAI, and AAAI.
\end{IEEEbiography}
\vspace{-0.5cm}
\begin{IEEEbiography}[{\includegraphics[width=1in,height=1.25in,clip,keepaspectratio]{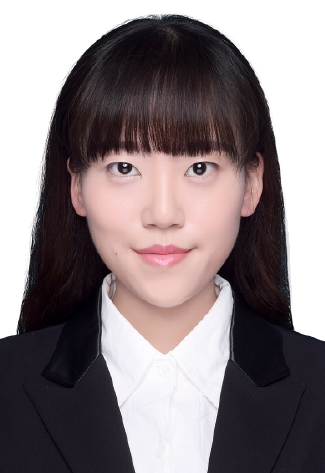}}]{Yumeng Xue} received the B.E. degree in communications engineering from Wuhan University, Wuhan, China, in 2015, and the M.S. degree in electronics and communications engineering from Xidian University, Xi'an, China, in 2018. Her research interests focus on hyperspectral image processing, pattern recognition, and deep learning.
\end{IEEEbiography}
\vspace{-0.5cm}
\begin{IEEEbiography}[{\includegraphics[width=1in,height=1.25in,clip,keepaspectratio]{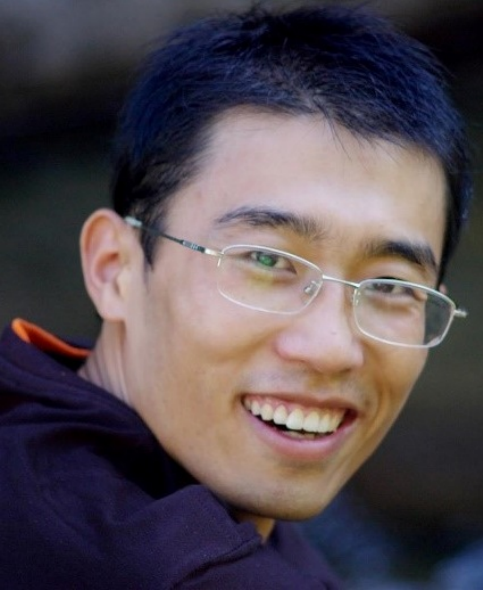}}]{Xianglong Liu} received the BS and Ph.D degrees in computer science from Beihang University, Beijing, in 2008 and 2014. From 2011 to 2012, he visited the Digital Video and Multimedia (DVMM) Lab, Columbia University as a joint Ph.D student. He is now an assistant professor with Beihang University. His research interests include machine learning, computer vision and multimedia information retrieval.
\end{IEEEbiography}
\vspace{-0.5cm}
\begin{IEEEbiography}[{\includegraphics[width=1in,height=1.25in,clip,keepaspectratio]{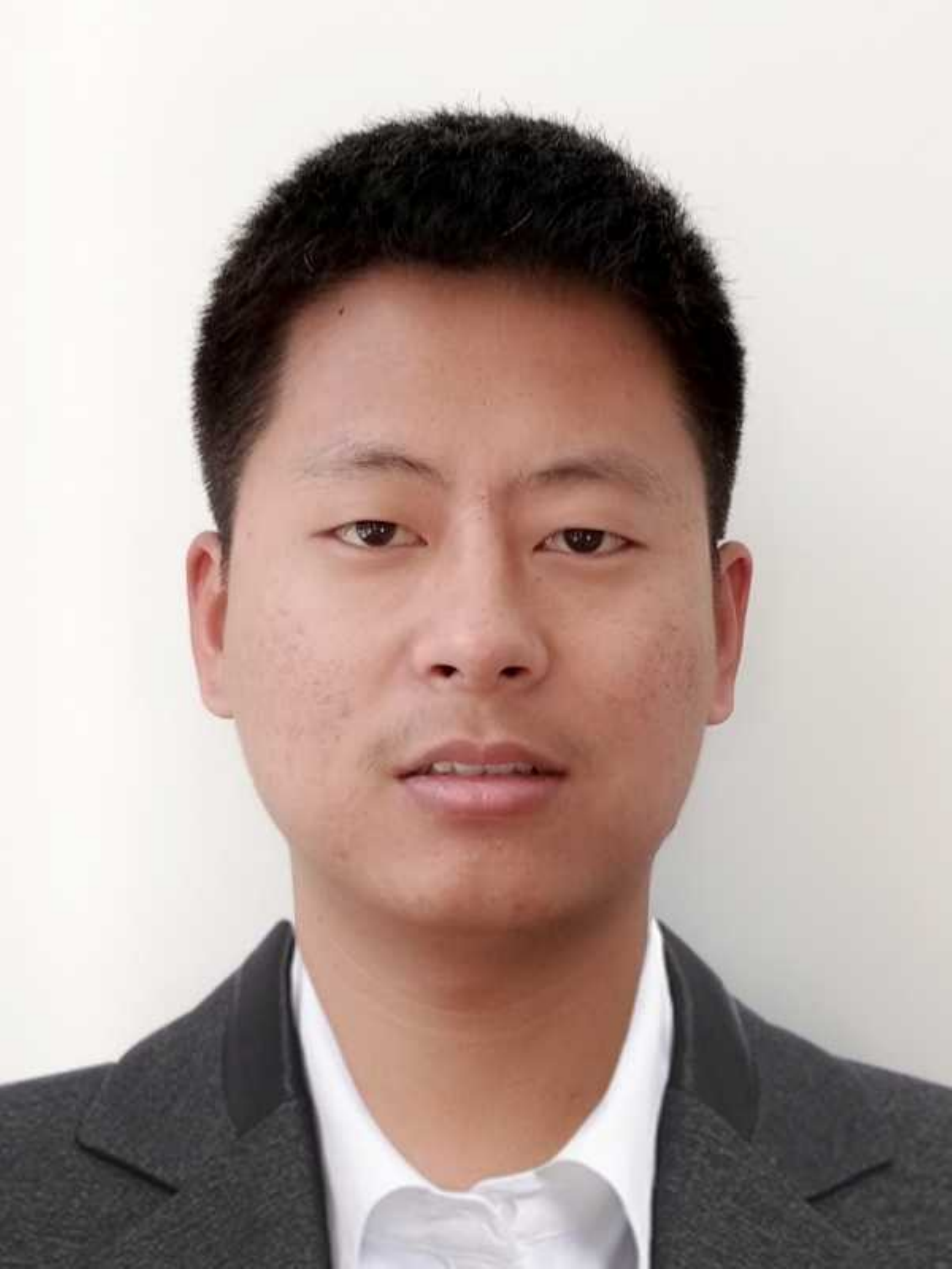}}]{Chao Li} received the B.E. degree in Electronic and Information Engineering from Inner Mongolia University of Science \& Technology, China, in 2014. He is currently pursuing his Ph.D. degree at School of Electronic Engineering, Xidian University. His main research interests include computer vision and machine learning.
\end{IEEEbiography}
\vspace{-0.5cm}
\begin{IEEEbiography}[{\includegraphics[width=1in,height=1.25in,clip,keepaspectratio]{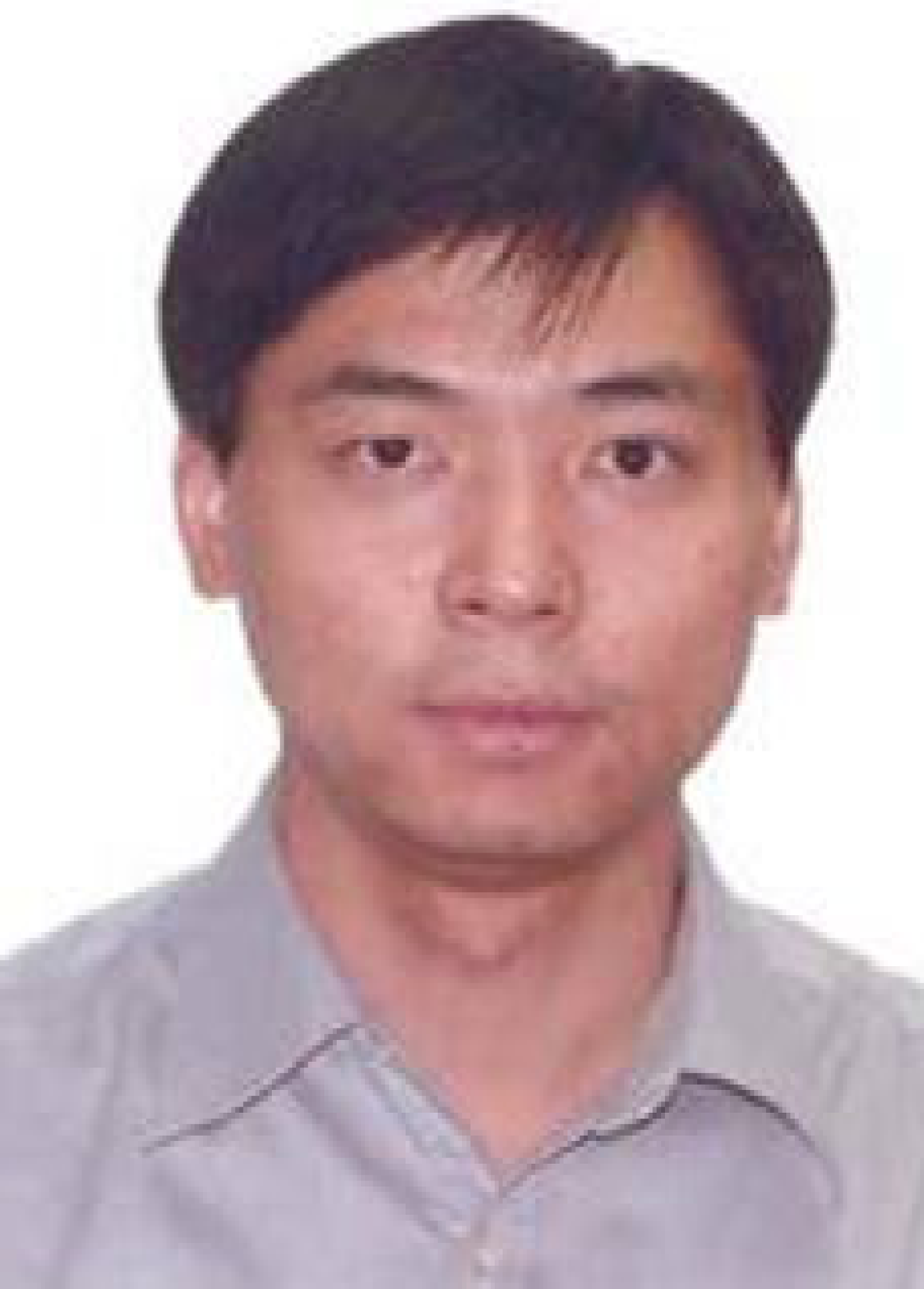}}]{Dacheng Tao} (F'15) is Professor of Computer Science and ARC Laureate Fellow in the School of Information Technologies and the Faculty of Engineering and Information Technologies, and the Inaugural Director of the UBTECH Sydney Artificial Intelligence Centre, at the University of Sydney. He mainly applies statistics and mathematics to Artificial Intelligence and Data Science. His research results have expounded in one monograph and 200+ publications at prestigious journals and prominent conferences, such as IEEE T-PAMI, T-IP, T-NNLS, IJCV, JMLR, NIPS, ICML, CVPR, ICCV, ECCV, ICDM; and ACM SIGKDD, with several best paper awards, such as the best theory/algorithm paper runner up award in IEEE ICDM’07, the best student paper award in IEEE ICDM’13, the distinguished paper award in the 2018 IJCAI, the 2014 ICDM 10-year highest-impact paper award, and the 2017 IEEE Signal Processing Society Best Paper Award. He is a Fellow of the Australian Academy of Science, AAAS, IEEE, IAPR, OSA and SPIE.
\end{IEEEbiography}

%
%




\end{document}